\documentclass[nohyperref]{article}

\usepackage{microtype} 
\usepackage{graphicx}
\usepackage{subfigure} 
\usepackage{booktabs} 

\usepackage{hyperref} 

\usepackage[accepted]{icml2022}

\usepackage{amsmath}
\usepackage{amssymb}
\usepackage{mathtools}
\usepackage{amsthm}
\usepackage{textcomp}

\newcommand{\x}{\mathbf{x}}
\newcommand{\z}{\mathbf{z}}

\def\Eqref#1{Eq.~\ref{#1}}

\theoremstyle{plain}
\newtheorem{theorem}{Theorem}
\newtheorem{corollary}{Corollary}

\newtheorem{definition}{Definition}
\newtheorem{proposition}{Proposition}
\newtheorem{remark}{Remark}

\usepackage{xcolor,colortbl}
\definecolor{ourmethod}{gray}{0.93}
\definecolor{myredcolor}{RGB}{215,48,39}
\definecolor{mygreencolor}{RGB}{26,152,80}
\usepackage{pifont}
\newcommand{\cmark}{\textcolor{black}{\ding{51}}}
\newcommand{\xmark}{\textcolor{black}{\ding{55}}}

\usepackage{enumitem}

\usepackage[textsize=tiny]{todonotes}

\icmltitlerunning{Learning Latent Causal Dynamics} 

\begin{document}

\twocolumn[
\icmltitle{Learning Latent Causal Dynamics}



\begin{icmlauthorlist}
\icmlauthor{Weiran Yao}{cmu}
\icmlauthor{Guangyi Chen}{cmu,mbzuai}
\icmlauthor{Kun Zhang}{cmu,mbzuai}
\end{icmlauthorlist}

\icmlaffiliation{cmu}{Carnegie Mellon University, Pittsburgh, PA, USA}
\icmlaffiliation{mbzuai}{Mohamed bin Zayed University of Artificial Intelligence (MBZUAI), Abu Dhabi, United Arab Emirates}

\icmlcorrespondingauthor{Kun Zhang}{kunz1@cmu.edu}

\icmlkeywords{Nonlinear ICA, Causal discovery, Latent variable model, State space model, Representation learning}

\vskip 0.3in
]



\printAffiliationsAndNotice{} 

\begin{abstract}
One critical challenge of time-series modeling is how to learn and quickly correct the model under unknown distribution shifts. In this work, we propose a principled framework, called \textit{LiLY}, to first recover time-delayed latent causal variables and identify their relations from measured temporal data under different distribution shifts. The correction step is then formulated as learning the low-dimensional change factors with a few samples from the new environment, leveraging the identified causal structure. Specifically, the framework factorizes unknown distribution shifts into transition distribution changes caused by fixed dynamics and time-varying latent causal relations, and by global changes in observation. We establish the identifiability theories of nonparametric latent causal dynamics from their nonlinear mixtures under fixed dynamics and under changes. Through experiments, we show that time-delayed latent causal influences are reliably identified from observed variables under different distribution changes. By exploiting this modular representation of changes, we can efficiently learn to correct the model under unknown distribution shifts with only a few samples. 

\end{abstract}

\section{Introduction} 
\begin{figure}[ht]
\centering
\includegraphics[width=\linewidth, height=8cm]{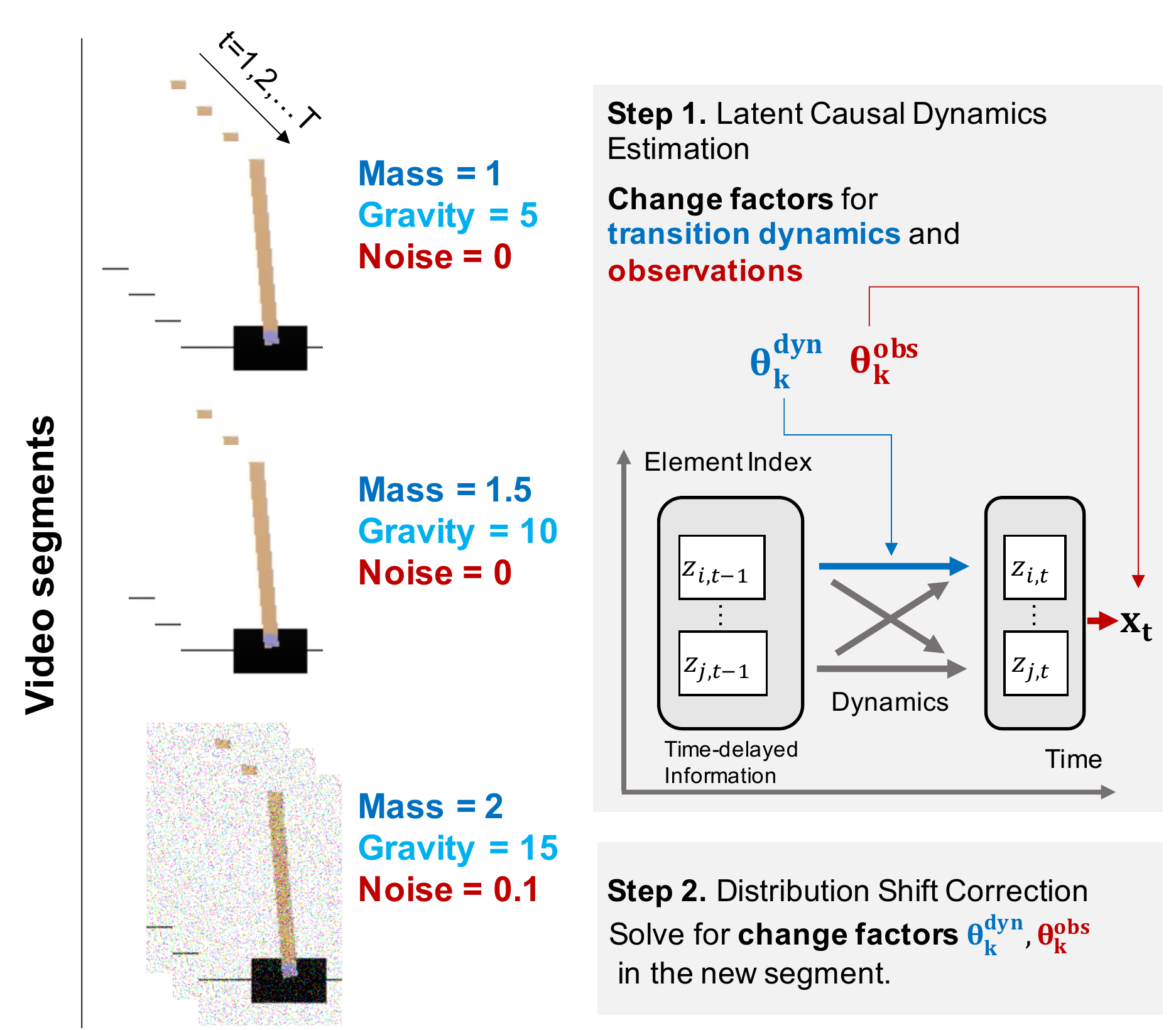}
\vspace{-0.2cm}
\caption{\textbf{LiLY}: \underline{L}earn\underline{i}ng \underline{L}atent causal d\underline{Y}namics under modular distribution shift. We exploit distribution changes resulting from fixed causal dynamics, changing causal influences and global observation changes to identify the underlying causal dynamics. The distribution change in a new segment is corrected via learning the low-dimensional change factors in an unsupervised way.}
\label{fig:problem}
\vspace{-0.2cm}
\end{figure}

Unknown distribution shifts or nonstationarity in natural signals have posed a critical challenge for time-series forecasting. For example, if the distribution shifts (e.g., changing dynamics, heterogeneous noise, etc.) are not well captured, models fit on these data samples will either overfit specific kinds of changes, or learn to use only the stationary parts for prediction, both of which significantly degrade the predictive performances.  One prevalent approach in the literature is to actively transfer the time-series model under the distribution shifts. However, if one is not aware of how the changes influence the underlying data generating process, one may not know how to target specific mechanisms and change the learned model efficiently. In other words, the true underlying processes or a graphical model that encode the change property of the distribution need to be given as a priori. To accomplish this step, causal discovery is often used. Causal discovery aims to identify the underlying structure of the data generation process by exploiting an appropriate class of assumptions \citep{SGS93,Pearl00}. Although distribution changes are not desired by time-series models, these changes, serving as a ``soft'' way of intervention, have proved to greatly improve the identifiability results for learning the latent causal structure \citep{, yao2021learning,bengio2019meta, ke2019learning}. For this reason, a natural solution  is to first exploit the distribution changes to identify the latent causal dynamics, and then use the uncovered structure to correct the model under changes.  

In this work, we propose a principled framework, called \textit{LiLY}, to first recover time-delayed latent causal variables and identify their relations from measured temporal data under different distribution shifts. The transfer step is then formulated as learning the low-dimensional change factors with a few samples from the new environment, leveraging the identified causal structure. For instance,  Fig.~\ref{fig:problem} shows an example of 
multiple video segments of a physical system (i.e., cartpole) under different mass, gravity and environment rendering settings. With LiLY, the differences across segments are characterized by change factors $\boldsymbol{\theta}_k^{\text{dyn}}$ of segment $k$ that encode changes in transition dynamics, and changes in observation modeled by $\boldsymbol{\theta}_k^{\text{obs}}$. We then present a generalized time-series data generative model that takes these change factors as arguments for modeling the distribution changes. Specifically, the proposed latent variable model factorizes distribution changes in fixed dynamics, time-varying transitions and observation by constructing partitioned latent subspaces. If the underlying latent causal dynamics augmented with change factors is uncovered, correcting distribution shifts can be performed efficiently by estimating the change factors in a new segment. 

\vspace{-0.5cm}
\begin{table}[h]
\centering
\begin{minipage}[b]{\linewidth}
\centering
\caption{Attributes of prior nonlinear ICA theories for time-series. A check denotes that a method has an attribute, whereas a cross denotes the opposite. $^\dagger$ indicates our approach.\label{tab:prior}}
\resizebox{\textwidth}{!}{
\begin{tabular}{lcccc}
\toprule
{\bf Theory} &
  \begin{tabular}[c]{@{}c@{}}\bf Time-varying \\ \bf Relation\end{tabular} &
  \begin{tabular}[c]{@{}c@{}}\bf Causally-related \\ \bf Process\end{tabular} &
  \begin{tabular}[c]{@{}c@{}}\bf Partitioned \\ \bf Subspace\end{tabular} &
  \begin{tabular}[c]{@{}c@{}}\bf Nonparametric\\ \bf Transition\end{tabular}  \\ \hline
{PCL} & \xmark & \xmark & \xmark & \cmark   \\
{HM-NLICA} & \xmark & \xmark & \xmark & \xmark   \\
{SlowVAE}  & \xmark & \xmark & \xmark & \xmark  \\
{SNICA} &\cmark & \cmark & \xmark &\xmark \\
{LEAP} &\xmark & \cmark & \xmark & \cmark \\
\rowcolor{ourmethod} {\textbf{LiLY} $^\dagger$} & \cmark & \cmark & \cmark & \cmark \\
\bottomrule
\end{tabular}
}
\end{minipage}
\end{table}

However, estimating latent causal structure from observations is notoriously challenging because the latent variables, even with independent factors of variation \citep{locatello2019challenging}, are not identifiable in the most general case \citep{hyvarinen1999nonlinear}. There exist several pieces of work aiming to uncover causally related latent variables under  linear relations or minimality assumptions. As early as \citep{Spearman28}, vanishing Tetrad conditions were exploited to identify latent variables in linear-Gaussian models \citep{Silva06}. Generalized Independent Noise (GIN) condition estimates linear, non-Gaussian latent variable causal graph \citep{xie2020generalized}. Recently, in the field of nonlinear Independent Component Analysis (ICA), identifiability results have been established \citep{hyvarinen2016unsupervised,hyvarinen2017nonlinear,hyvarinen2019nonlinear,khemakhem2020variational,sorrenson2020disentanglement} by using certain side information, such as class labels, in addition to independence. For time-series data, history information is widely used as side information for nonlinear ICA. However, most existing work that establishes identifiability results considers either stationary independent sources such as PCL \citep{hyvarinen2017nonlinear}, SlowVAE \citep{klindt2020towards} or under linear transition assumptions such as SlowVAE \citep{klindt2020towards} and SNICA \citep{halva2021disentangling}, or with certain structure such as Markov properties in HM-NLICA \citep{halva2020hidden}. LEAP \citep{yao2021learning}, which is the closest work to ours,  has established the identifiability of the nonparametric latent temporal processes in certain nonstationary cases,  under the condition that the distribution of the noise terms of the latent processes vary across segments. 

Our theory can be differentiated from the prior work in four aspects as summarized in Table~\ref{tab:prior}:

\begin{enumerate}[leftmargin=*] \itemsep0em
    \item We consider a very general nonstationary case which includes changing causal dynamics such as changes in the influencing strength or switching some edges off \citep{li2020causal}, nonstationary noise distributions across domains \citep{yao2021learning}, and domain shifts in observation \citep{sahoo2021contrast}, etc., as special cases; 
    \item We establish the identifiability results for nonparametric, causally-related time-delayed latent processes, instead of independent processes, or under linear assumptions;
    \item We allow both stationary and nonstationary latent processes to co-exist by using partitioned latent subspaces in the data generating processes. The identifiability results under this setting are appealing to real-world data where whether and how the distribution changes are unknown. Our identifiability condition is generally much weaker than the ones in \citep{yao2021learning} because we further benefit from distribution changes under fixed causal dynamics, while LEAP assumes all latent processes are changed across contexts.
\end{enumerate}

Through experiments, we show that time-delayed latent causal processes are reliably identified from observed variables under different distribution shifts and dependency structures. By exploiting this modular representation of changes, we can efficiently learn to correct the model under unknown distribution shifts with only a few samples.

\section{Related Work}  

\subsection{Nonlinear ICA for Time Series}
Temporal structure and nonstationarities were used to achieve identifiability of nonlinear ICA. \citet{hyvarinen2016unsupervised} proposed time-contrastive learning (TCL) based on the independent sources assumption and leverage variability in variance terms. \cite{hyvarinen2017nonlinear} developed a permutation-based contrastive (PCL) learning framework which discriminates between true independent sources and permuted sources, and the model is identifiable under the uniformly dependent assumption. \cite{halva2020hidden} combined nonlinear ICA with a Hidden Markov Model (HMM) to automatically model nonstationarity without the need for manual data segmentation. \cite{khemakhem2020variational} introduced VAEs to approximate the true joint distribution over observed and auxiliary nonstationary regimes. The conditional distribution in their work is assumed to be within exponential families to achieve identifiability on the latent space. The most recent literature on nonlinear ICA for time-series is the work of \citep{yao2021learning}, which proposed both a nonparametric condition leveraging the nonstionary noise terms, and a linear, parametric condition leveraging the functional form with generalized Laplacian properties of the noise terms. 

\subsection{Causal Discovery from Time Series}

Inferring the causal structure from time-series data is critical to many fields including machine learning~\citep{berzuini2012causality},  econometrics~\citep{ghysels2016testing}, and neuroscience~\citep{friston2009causal}. Most existing work focuses on estimating the temporal causal relations between observed variables. For this task, constraint-based methods~\citep{entner2010causal} apply the conditional independence tests to recover the causal structures, while score-based methods~\citep{murphy2002dynamic,pamfil2020dynotears}
define score functions to guide a search process. Furthermore, 
~\citet{malinsky2018causal,malinsky2019learning} propose to fuse both conditional independence tests and score-based methods. The Granger causality~\citep{granger1969investigating} and its non-linear variations~\citep{tank2018neural, lowe2020amortized} are also widely used. 

\section{Problem Formulation} 

We first present a holistic, modular time-series data generative model that factorizes distribution changes in fixed dynamics, time-varying transitions and observation by constructing partitioned latent subspaces. We then define the properties of the generative model which suffice the identifiability of time-delayed causal dynamics. Finally, we formulate the problem of  distribution shift correction as learning the change factor representation in a new environment with the identified latent causal structure. 

\subsection{A Modular Representation of Changes}

Following \citep{huang2020causal, huang2021adarl}, we can represent the changes in a compact way by using a low-dimensional vector $\boldsymbol{\theta}_k$. Specifically, let $\mathbf{u}$ denotes domain or segment index. Suppose there exist $K$ domains or segments, i.e., $u_k$ with $k=1,2,...,K$. In each segment, we observe time series $\{\mathbf{x}_t\}_{t=1}^T$ generated by an arbitrary invertible nonlinear mixture of the underlying temporal processes $\mathbf{z}_t$.  We partition the latent space into three blocks $\mathbf{z}_t = (\mathbf{z}_t^{\text{fix}}, \mathbf{z}_t^{\text{chg}}, \mathbf{z}_t^{\text{obs}})$ where $z_{s,t}^{\text{fix}}$ denotes the s\textsuperscript{th} component of the fixed dynamics parts, $z_{c,t}^{\text{chg}}$ denotes the c\textsuperscript{th} component of the changing dynamics parts, and $\mathbf{z}_{o,t}^{\text{obs}}$ is the o\textsuperscript{th} component of the global changes (e.g., styles). We assume the data generating process in each segment $\mathbf{u}_k$ can be described by the transition function $f$ for each dimension of $\mathbf{z}_t$ that takes its parent nodes change factors, $(\boldsymbol{\theta}_k^{\text{dyn}}, \boldsymbol{\theta}_k^{\text{obs}})$ and noise terms as inputs, and the mixing function $\mathbf{g}$ as: 

\vspace{-0.1cm}
\begin{equation}
 \small
 \vspace{-0.2cm}
 \setlength{\abovedisplayskip}{4pt}
 \setlength{\belowdisplayskip}{4pt}
\left \{
 \begin{array}{lll}
     z_{s, t}^{\text{fix}} &= f_s\left(\{z_{i, t-\tau} \vert z_{i, t-\tau} \in \mathbf{Pa}(z_{s,t}^{\text{fix}}) \}, \epsilon_{s,t}  \right), \\
     z_{c, t}^{\text{chg}} &= f_c\left(\{z_{i, t-\tau} \vert z_{i, t-\tau} \in \mathbf{Pa}(z_{c,t}^{\text{chg}}) \}, \boldsymbol{\theta}_k^{\text{dyn}}, \epsilon_{c,t}  \right), \\
     z_{o, t}^{\text{obs}} &= f_o\left(\boldsymbol{\theta}_k^{\text{obs}}, \epsilon_{o,t}  \right), \\
    \mathbf{x}_t &= \mathbf{g}(\mathbf{z}_t),
 \end{array}\right.
 \label{eq:model}
\end{equation} 

where $\mathbf{Pa}(z_{i,t})$ denote the set of (time-delayed) parent nodes of $z_{i,t}$, and $\epsilon_{s,t}, \epsilon_{c,t}, \epsilon_{o,t}$ terms are mutually independent (i.e., spatially and temporally independent) random noises in each segment. The vector $\boldsymbol{\theta}_k = (\boldsymbol{\theta}_k^{\text{dyn}}, \boldsymbol{\theta}_k^{\text{obs}})$ embeds the information of segment $u_k$.  Note that with Eq.~\ref{eq:model}, we allow a very general class of changes in the model, which includes changing causal dynamics such as changes in the influencing strength or switching some edges off \citep{li2020causal}, nonstationary noise distributions across segments \citep{yao2021learning}, and domain shifts in observation \citep{sahoo2021contrast}, etc., all as special cases.  

\begin{remark}

The mutual independence of noise terms over space and time imply that: (1) $\epsilon_{i, t}$ are independent from the parent nodes $\mathbf{Pa}(z_{i,t})$, and (2) $z_{i,t}$ are conditional independent given the history information $\mathbf{z}_{\text{Hx}} = \{\mathbf{z}_{t-\tau}\}_{\tau=1}^{L}$ up to maximum time lag $L$  and segment embeddings or index.

\end{remark}

This Remark suggests that the Independent Noise (IN) condition \citep{pearl2000models} and Conditional Independence (CI) condition \citep{hyvarinen2019nonlinear}, both of which are common assumptions for causal discovery, are naturally satisfied in time-series data under random noise. 

\subsection{Identifiability of Time-Delayed Causal Dynamics}

We define the identifiability of time-delayed latent causal dynamics in the representation function space. If the latent variables can be identified at least up to permutation and component-wise invertible nonlinearities, we say that latent causal dynamics are also identifiable because conditional independence relations fully characterize time-delayed causal relations in a time-delayed causally sufficient system.

\begin{definition}[Identifiable Latent Causal Dynamics]
Formally let $\{\mathbf{x}_t\}_{t=1}^T$ be a sequence of observed variables generated by the true temporally causal latent processes specified by $(f_i, \boldsymbol{\theta}_k, p({\epsilon_i}), \mathbf{g})$ given in Eq.~\ref{eq:model}. A learned generative model $(\hat{f}_i, \hat{\boldsymbol{\theta}}_k, \hat{p}({\epsilon_i}), \hat{\mathbf{g}})$ is observationally equivalent to $(f_i, \boldsymbol{\theta}_k, p({\epsilon_i}), \mathbf{g})$ if the model distribution $p_{\hat{f}, \hat{\boldsymbol{\theta}}_k, \hat{p}_\epsilon, \hat{\mathbf{g}}}(\{\mathbf{x}_t\}_{t=1}^T)$ matches the data distribution $ p_{f, \boldsymbol{\theta}_k, p_\epsilon, \mathbf{g}}(\{\mathbf{x}_t\}_{t=1}^T)$ everywhere. We say latent causal processes are identifiable if observational equivalence can lead to identifiability of the latent variables up to permutation $\pi$ and component-wise invertible transformation $T$:
\begin{equation}\label{eq:iden}
\setlength{\abovedisplayskip}{4pt}
\setlength{\belowdisplayskip}{4pt}
\begin{aligned}
p_{\hat{f}_i, \hat{\boldsymbol{\theta}}_k, \hat{p}_{\epsilon_i}, \hat{\mathbf{g}}}(&\{\mathbf{x}_t\}_{t=1}^T) =  p_{f_i, \boldsymbol{\theta}_k, p_{\epsilon_i}, \mathbf{g}} (\{\mathbf{x}_t\}_{t=1}^T)  \\
\Rightarrow \hat{\mathbf{g}} &(\mathbf{x}_t) = \mathbf{g} \circ \pi \circ T, \quad \forall \mathbf{x}_t \in \mathcal{X},
\end{aligned}
\end{equation}
where $\mathcal{X}$ is the observation space.
\end{definition} 

\subsection{Distribution Shift Correction}

We first assume that the latent causal dynamics model in Eq.~\ref{eq:model} is identified (we will explain under what conditions the identifiability can be achieved in Sec.~\ref{sec:theory} and how in Sec.~\ref{sec:method}). Then correcting distribution shifts is equivalent to learning the $\boldsymbol{\theta}_k$ embeddings in the new environment. In practice, distributions are often changed in a sparse or local way given the causal disentanglement, i.e, only a few factors may change simultaneously, which is known as sparse mechanism shift assumption \citep{scholkopf2021toward}. Following the assumption, we encode $\boldsymbol{\theta}_k$ with low-dimensional embeddings, in order to capture only the domain differences. 

\section{Theory}\label{sec:theory} 

In this section, we establish the identifiability theory of nonparametric time-delayed latent causal processes under three different types of distribution shifts. W.l.o.g., we consider the latent processes with maximum time lag $L=1$. In particular, \textbf{(1)} under fixed causal dynamics, we leverage the distribution changes $p(z_{k,t} \vert \mathbf{z}_{t-1})$ for different values of $\mathbf{z}_{t-1}$; \textbf{(2)} under changing causal dynamics, we exploit the changing causal influences on $p(z_{k,t} \vert \mathbf{z}_{t-1}, u_k)$ under different domain $u_k$, and \textbf{(3)} under global observation changes, the nonstationarity $p(z_{k,t} \vert u_k)$ under different values of $u_k$ is exploited. We illustrate through examples that, even without explicit changes in causal influences, the latent causal processes are generally identifiable if the process noises are not perfectly Gaussian. We show that the identifiability results can further benefit from nonstationarity with weaker identifiability conditions. This identifiability theory extends the prior work \citep{yao2021learning, klindt2020towards, hyvarinen2017nonlinear} by allowing causal relations between latent factors, instead of independent sources, and partitioned subspaces with both fixed and/or changing causal influences. The proofs are provided in Appendix~\ref{ap:theory}. 

\subsection{Identifiability under Fixed Causal Dynamics}
Let $\eta_{kt} \triangleq \log p(z_{k, t} | \mathbf{z}_{t-1})$. Assume that $\eta_{kt}$ is twice differentiable in $z_{k,t}$ and is differentiable in $z_{l,t-1}$, $l=1,2,...,n$. Note that the parents of $z_{k,t}$ may be only a subset of $\mathbf{z}_{t-1}$; if $z_{l,t-1}$ is not a parent of $z_{k,t}$, then $\frac{\partial \eta_{kt}}{\partial z_{l,t-1}} = 0$.
Below we provide a {\it sufficient condition} for the identifiability of $\mathbf{z}_{t}$, followed by a discussion of specific unidentifiable and identifiable cases to illustrate how general it is.
 
\begin{theorem}[Identifiablity under a Fixed Temporal Causal Model] \label{Th1} 
Suppose there exists invertible function $\hat{\mathbf{g}}$  that maps $\mathbf{x}_t$ to $\hat{\mathbf{z}}_t$, i.e., 
\begin{equation} \label{eq:invert}
    \hat{\mathbf{z}}_t = \hat{\mathbf{g}}(\mathbf{x}_t)
\end{equation}
such that the components of $\hat{\mathbf{z}}_t$ are mutually  independent conditional on $\hat{\mathbf{z}}_{t-1}$. 
Let 
\begin{equation} \label{eq:v}
\begin{aligned}
\mathbf{v}_{k,t} &\triangleq \Big(\frac{\partial^2 \eta_{kt}}{\partial z_{k,t} \partial z_{1,t-1}}, \frac{\partial^2 \eta_{kt}}{\partial z_{k,t} \partial z_{2,t-1}}, ..., \frac{\partial^2 \eta_{kt}}{\partial z_{k,t} \partial z_{n,t-1}} \Big)^\intercal \\ \mathring{\mathbf{v}}_{k,t} &\triangleq \Big(\frac{\partial^3 \eta_{kt}}{\partial z_{k,t}^2 \partial z_{1,t-1}}, \frac{\partial^3 \eta_{kt}}{\partial z_{k,t}^2 \partial z_{2,t-1}}, ..., \frac{\partial^3 \eta_{kt}}{\partial z_{k,t}^2 \partial z_{n,t-1}} \Big)^\intercal. 
\end{aligned}
\end{equation}
If for each value of $\mathbf{z}_t$, $\mathbf{v}_{1,t}, \mathring{\mathbf{v}}_{1,t}, \mathbf{v}_{2,t}, \mathring{\mathbf{v}}_{2,t}, ..., \mathbf{v}_{n,t}, \mathring{\mathbf{v}}_{n,t}$, as 
$2n$ vector functions in $z_{1,t-1}$, $z_{2,t-1}$, ..., $z_{n,t-1}$, are linearly independent, then $\mathbf{z}_{t}$ must be an invertible, component-wise transformation of a permuted version of $\hat{\mathbf{z}}_t$.

\end{theorem}

The linear independence condition in Theorem \ref{Th1} is the core condition to guarantee the identifiability of $\mathbf{z}_t$ from the observed $\mathbf{x}_t$. To make this condition more intuitive, below we consider specific unidentifiable cases, in which there is no temporal dependence in $\mathbf{z}_t$ or the noise terms in $\mathbf{z}_t$ are additive Gaussian, and two identifiable cases, in which $\mathbf{z}_t$ has additive, heterogeneous noise or follows some linear, non-Gaussian temporal process. 

Let us start with two unidentifiable cases. In case $\texttt{N}1$, $\mathbf{t}_t$ is an independent and identically distributed (i.i.d.) process, i.e., there is no causal influence from any component of $\mathbf{z}_{t-1}$ to any $z_{k,t}$. In this case,  $\mathbf{v}_{k,t}$ and $\mathring{\mathbf{v}}_{k,t}$ (defined in Eq.~\ref{eq:v}) are always $\mathbf{0}$ for $k=1.,2,...,n$, since $p(z_{k,t} \,|\, \mathbf{z}_{t-1})$ does not involve $\mathbf{z}_{t-1}$. So th linear independence condition is violated. In fact, this is the regular nonlinear ICA problem with i.i.d. data, and it is well-known that the underlying independent variables are not identifiable \cite{hyvarinen1999nonlinear}.

In case $\texttt{N}_2$, all $z_{k,t}$ follow an additive noise model with Gaussian noise terms, i.e., 
\begin{equation} \label{eq:Gaussian_case}
    \mathbf{z}_t = \mathbf{q}(\mathbf{z}_{t-1}) + \mathbf{\epsilon}_t,
\end{equation}
where $\mathbf{q}$ is a transformation and the components of the Gaussian vector $\mathbf{\epsilon}_t$ are independent and also independent from $\mathbf{z}_{t-1}$. Then  $\frac{\partial^2 \eta_{kt}}{\partial z_{k,t}^2}$ is constant, and $\frac{\partial^3 \eta_{kt}}{\partial z_{k,t}^2 \partial z_{l,t-1}} \equiv 0$, violating the linear independence condition. In the following proposition we give some alternative solutions and verify the unidentifiability in this case.

\begin{proposition}[Unidentifiability under Gaussian Noise]

Suppose $\mathbf{x}_t = \mathbf{g}(\mathbf{z}_t)$ was generated by Eq.~\ref{eq:Gaussian_case}, where the components of $\mathbf{\epsilon}_t$ are mutually independent Gaussian and also independent from $\mathbf{z}_{t-1}$. Then any $\hat{\mathbf{z}}_t = \mathbf{D}_1 \mathbf{U} \mathbf{D}_2 \cdot {\mathbf{z}}_t$, where $\mathbf{D}_1$ is an arbitrary non-singular diagonal matrix, $\mathbf{U}$ is an arbitrary orthogonal matrix, and $\mathbf{D}_2$ is a diagonal matrix with $\mathbb{V}ar^{-1/2}(\epsilon_{k,t})$ as its $k^{\text{th}}$  diagonal entry, is a valid solution to satisfy the condition that the components of $\hat{\mathbf{z}}_t$ are mutually independent conditional on $\hat{\mathbf{z}}_{t-1}$.

\end{proposition}

Roughly speaking, for a randomly chosen conditional density function $p(z_{k,t}\,|\,\mathbf{z}_{t-1})$ in which $z_{k,t}$ is not independent from $\mathbf{z}_{t-1}$ (i.e., there is temporal dependence in the latent processes) and which does not follow an additive noise model with Gaussian noise, the chance for its specific second- and third-order partial derivatives to be linearly dependent is slim.
Now let us consider two cases in which the latent temporally processes $\mathbf{z}_t$ are naturally identifiable under some technical conditions. First consider case $\texttt{Y}1$, where $z_{k,t}$ follows a heterogeneous noise process, in which the noise variance depends on its parents:
\begin{equation} \label{eq:heteo}
    z_{k,t} = q_k(\mathbf{z}_{t-1}) + \frac{1}{b_k(\mathbf{z}_{t-1})}\epsilon_{k,t}.
\end{equation}
Here we assume $\epsilon_{k,t}$ is standard Gaussian and $\epsilon_{1,t}, \epsilon_{2,t}, .., \epsilon_{n,t}$ are mutually independent and independent from $\mathbf{z}_{t-1}$. $\frac{1}{b_k}$, which depends on $\mathbf{z}_{t-1}$, is the standard deviation of the noise in $z_{k,t}$. (For conciseness, we drop the argument of $b_k$ and $q_k$ when there is no confusion.) Note that in this model, if $q_k$ is 0 for all $k=1,2,...,n$, it reduces to a multiplicative noise model. 
The identifiability result of $\mathbf{z}_t$ is established in the following proposition.

\begin{corollary}[Identifiability under Heterogeneous Noise]

Suppose $\mathbf{x}_t = \mathbf{g}(\mathbf{z}_t)$ was generated according to Eq.~\ref{eq:heteo}, and Eq.~\ref{eq:invert} holds true. If  $b_k\cdot \frac{\partial b_k}{\partial \mathbf{z}_{t-1}}$ and $b_k\cdot \frac{\partial b_k}{\partial \mathbf{z}_{t-1}} (z_{k,t} - q_{k}) - b_k^2\cdot \frac{\partial q_{k}}{\partial \mathbf{z}_{t-1}}$, with $k=1,2,...,n$, which are in total $2n$ function vectors in $\mathbf{z}_{t-1}$, 
are linearly independent, then $\mathbf{z}_{t}$ must be an invertible, component-wise transformation of a permuted version of $\hat{\mathbf{z}}_t$.

\end{corollary}

Let us then consider another special case, denoted by $\texttt{Y}2$, with linear, non-Gaussian temporal model for $\mathbf{z}_t$: the latent processes follow Eq.~\ref{eq:Gaussian_case}, with $\mathbf{q}$ being a linear transformation and $\epsilon_{k,t}$ following a particular class of non-Gaussian distributions. 
The following corollary shows that $\mathbf{z}_t$ is identifiable as long as each $z_{k,t}$ receives causal influences from some components of $\mathbf{z}_{t-1}$.

\begin{corollary}[Identifiability under a Specific Linear, Non-Gaussian Model for Latent Processes]

Suppose $\mathbf{x}_t = \mathbf{g}(\mathbf{z}_t)$ was generated according to Eq.~\ref{eq:Gaussian_case}, in which $\mathbf{q}$ is a linear transformation and for each $z_{k,t}$, there exists at least one $k'$ such that $c_{k,k'} \triangleq \frac{\partial z_{k,t}}{\partial z_{k',t-1}} \neq 0$. Assume the noise term $\epsilon_{k,t}$ follows a zero-mean generalized normal distribution:
\begin{equation}
p(\epsilon_{k,t}) \propto e^{-\lambda |\epsilon_{k,t}|^\beta}\textrm{,~~with positive $\lambda$ and }\beta > 2 \textrm{ and } \beta \neq 3.
\end{equation}
If Eq.~\ref{eq:invert} holds, then $\mathbf{z}_{t}$ must be an invertible, component-wise transformation of a permuted version of $\hat{\mathbf{z}}_t$.

\end{corollary}

\subsection{Further Benefits from Changing Causal Influences}

\citet{yao2021learning} established the identifiability of the latent temporal causal processes $\mathbf{z}_t$ in certain nonstationary cases, under the condition that the noise term in each $z_{k,t}$, relative to its parents in $\mathbf{z}_{t-1}$, changes across $m$ contexts corresponding to $\mathbf{u} = u_1, u_2, ..., u_m$. Here we show that the identifiability result shown in the previous section can further benefit from nonstationarity of the causal model, and our identifiability condition is generally much weaker than that in \citep{yao2021learning}: we allow changes in the noise term or causal influence on $z_{k,t}$ from its parents in $\mathbf{z}_{t-1}$, and our ``sufficient variability" condition is just a necessary condition for that in \citep{yao2021learning} because of the additional information that one can leverage.
Let $\mathbf{v}_{k,t}(u_r)$ be $\mathbf{v}_{k,t}$, which is defined in Eq.~\ref{eq:v}, in the $u_r$ context. Similarly, Let $\mathring{\mathbf{v}}_{k,t}(u_r)$ be $\mathring{\mathbf{v}}_{k,t}$ in the $u_r$ context. Let 
 $$\mathbf{s}_{k,t} \triangleq \Big( \mathbf{v}_{k,t}(u_1)^\intercal, ..., 
\mathbf{v}_{k,t}(u_m)^\intercal, 
\Delta^2_{2}, ...,
 \Delta^2_{m}
 \Big)^\intercal,$$
 $$\mathring{\mathbf{s}}_{k,t} \triangleq \Big( \mathring{\mathbf{v}}_{k,t}(u_1)^\intercal, ..., 
\mathring{\mathbf{v}}_{k,t}(u_m)^\intercal, 
\Delta_{2}, ...,
 \Delta_{m}
 \Big)^\intercal,$$
 where  
$\Delta^2_{i}= \frac{\partial^2 \eta_{kt}({u}_{i})}{\partial z_{k,t}^2 } - 
 \frac{\partial^2 \eta_{kt}({u}_{i-1})}{\partial z_{k,t}^2 } $ and $\Delta_{i}= \frac{\partial \eta_{kt}({u}_{i})}{\partial z_{k,t} } - 
 \frac{\partial \eta_{kt}({u}_{i-1})}{\partial z_{k,t} }. $ 
As provided below, in our case, the identifiablity of $\mathbf{z}_t$ is guaranteed by the linear independence of the whole function vectors $\mathbf{s}_{k,t}$ and $\mathring{\mathbf{s}}_{k,t}$, with $k=1,2,...,n$.
However, the identifiability result in \citep{yao2021learning} relies on the linear independence of only the last $m-1$ components of $\mathbf{s}_{k,t}$ and $\mathring{\mathbf{s}}_{k,t}$ with $k=1,2,...,n$; this linear independence is generally a much stronger condition.

\begin{theorem}[Identifiability under Changing Causal Dynamics]

Suppose $\mathbf{x}_t = \mathbf{g}(\mathbf{z}_t)$ and that the conditional distribution $p(z_{k,t} \,|\, \mathbf{z}_{t-1})$ may change across $m$ values of the context variable $\mathbf{u}$, denoted by $u_1$, $u_2$, ..., $u_m$. Suppose the components of $\mathbf{z}_t$ are mutually independent conditional on $\mathbf{z}_{-1}$ in each context. Assume that the components of $\hat{\mathbf{z}}_t$ produced by Eq.~\ref{eq:invert} are also mutually independent conditional on $\hat{\mathbf{z}}_{t-1}$. 
If the $2n$ function vectors $\mathbf{s}_{k,t}$ and $\mathring{\mathbf{s}}_{k,t}$, with $k=1,2,...,n$, are linearly independent, then $\hat{\mathbf{z}}_t$ is a permuted invertible component-wise transformation of $\mathbf{z}_t$. 
\end{theorem}

\begin{figure*}[ht]
\centering
\includegraphics[width=0.95\textwidth]{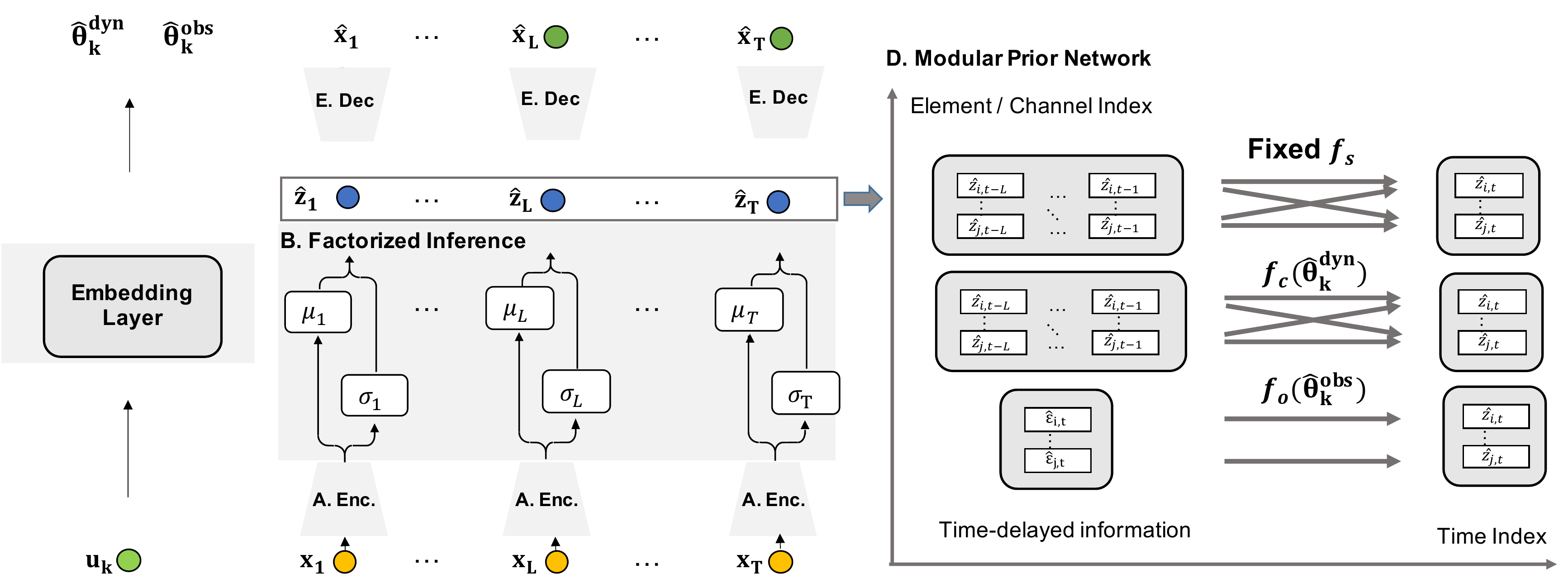}
\vspace{-0.2cm}
\caption{\textbf{Framework overview:} Encoder (A) and Decoder (E) using MLP or CNN for specific data types; (B) Factorized inference network that approximates the posteriors of latent variables $\mathbf{\hat{z}}_t$, (C) Change factor representation learning with embedding layer, and (D) Modular prior  network that factorizes distribution changes, i.e., fixed causal dynamics, changing causal dynamics, and global changes, into three partitioned latent causal processes constraint (Thm 1,2 and 3).}
\label{fig:arch}
\vspace{-0.2cm}
\end{figure*}

\begin{theorem}[Identifiability under Observation Changes]
Suppose $\mathbf{x}_t = \mathbf{g}(\mathbf{z}_t)$ and that the conditional distribution $p(z_{k,t} \,|\, \mathbf{u})$ may change across $m$ values of the context variable $\mathbf{u}$, denoted by $u_1$, $u_2$, ..., $u_m$. Suppose the components of $\mathbf{z}_t$ are mutually independent conditional on $\mathbf{u}$ in each context. Assume that the components of $\hat{\mathbf{z}}_t$ produced by Eq.~\ref{eq:invert} are also mutually independent conditional on $\hat{\mathbf{z}}_{t-1}$. 
If the $2n$ function vectors $\mathbf{s}_{k,t}$ and $\mathring{\mathbf{s}}_{k,t}$, with $k=1,2,...,n$, are linearly independent, then $\hat{\mathbf{z}}_t$ is a permuted invertible component-wise transformation of $\mathbf{z}_t$.

\end{theorem}

\begin{corollary}[Identifiability under Modular Distribution Shifts] Assume the data generating process in Eq.~\ref{eq:model}. If the three partitioned latent components $\mathbf{z}_t = (\mathbf{z}_t^{\text{fix}}, \mathbf{z}_t^{\text{chg}}, \mathbf{z}_t^{\text{obs}})$ respectively satisfy the conditions in \textbf{Theorem} 1, \textbf{Theorem} 2, and \textbf{Theorem} 3, then $\mathbf{z}_{t}$ must be an invertible, component-wise transformation of a permuted version of $\hat{\mathbf{z}}_t$. 
 
\end{corollary}

\section{LiLY: Learning Latent Causal Dynamics}\label{sec:method}

Given our identifiability results, we propose \underline{L}earn\underline{i}ng \underline{L}atent causal d\underline{Y}namics (LiLY)  framework to estimate the latent causal dynamics under modular distribution shifts, by extending Sequential Variational Auto-Encoders \cite{li2018disentangled} with tailored modules to model different distribution shifts, and enforcing the conditions
in Sec.~\ref{sec:theory}  as constraints. We give the estimation procedure of the latent causal dynamics model in Eq.~\ref{eq:model}. The model architecture is showcased in Fig.~\ref{fig:arch}. The framework has the following three major components. The implementation details are in Appendix~\ref{ap:arch}. 

\paragraph{Modular Prior Network} The core component of the estimation is to learn the three types of distribution changes. We leverage the partitioned estimated latent subspaces $\hat{\mathbf{z}}_t = (\hat{\mathbf{z}}_t^{\text{fix}}, \hat{\mathbf{z}}_t^{\text{chg}}, \hat{\mathbf{z}}_t^{\text{obs}})$ and model their distribution changes in conditional priors. In particular, for (1) fixed causal dynamics processes $\hat{\mathbf{z}}_t^{\text{fix}}$, their transition priors  are obtained by first learning inverse transition functions $f_s^{-1}$ that take the estimated latent variables and output random noise terms, and applying the change of variables formula to the transformation: $p(\hat{z}_{s,t}^{\text{fix}} \vert \hat{\mathbf{z}}_{\text{Hx}}) = p_{\epsilon_s}\left(f_s^{-1}(\hat{z}_{s,t}^{\text{fix}}, \hat{\mathbf{z}}_{\text{Hx}})]\right)\Big|\frac{\partial f_s^{-1}}{\partial \hat{z}_{s,t}^{\text{fix}}}\Big|$; (2) for changing causal dynamics, we evaluate $p(\hat{z}_{s,t}^{\text{chg}} \vert \hat{\mathbf{z}}_{\text{Hx}}, \mathbf{u}_k) = p_{\epsilon_c}\left(f_c^{-1}(\hat{z}_{c,t}^{\text{chg}}, \hat{\mathbf{z}}_{\text{Hx}}, \hat{\boldsymbol{\theta}}_k^{\text{dyn}})]\right)\Big|\frac{\partial f_c^{-1}}{\partial \hat{z}_{c,t}^{\text{chg}}}\Big|$ by learning a holistic inverse dynamics $f_c^{-1}$ that takes the estimated change factors for dynamics $\hat{\boldsymbol{\theta}}_k^{\text{dyn}})$ as inputs, and similarly for (3) observation changes $\hat{\mathbf{z}}_t^{\text{obs}}$, we learn to project them to invariant noise terms by $f_o^{-1}$ which takes the change factors $\boldsymbol{\theta}_k^{\text{obs}}$ as arguments, and obtains  $p(\hat{z}_{o,t}^{\text{obs}} \vert \mathbf{u}_k) = p_{\epsilon_o}\left(f_o^{-1}(\hat{z}_{o,t}^{\text{obs}}, \hat{\boldsymbol{\theta}}_k^{\text{obs}})]\right)\Big|\frac{\partial f_o^{-1}}{\partial \hat{z}_{o,t}^{\text{obs}}}\Big|$ as the prior. Two immediate benefits are brought by learning the inverse dynamics function: (1) compared with  forward prediction, learning to recover the noise is more appealing for nonparametric processes, in the sense that using forward prediction with a fixed loss function, one cannot model latent processes without parametric forms. For example, heterogeneous noise model in the latent processes (Eq.~\ref{eq:heteo}) cannot be estimated by forward prediction with $L_2$ loss, and (2) (conditional) mutual independence of the estimated latent variables $\hat{\mathbf{z}}_t$ can be easily enforced by summing up all estimated component densities when obtaining the joint $p(\mathbf{z}_t \vert \mathbf{z}_{\text{Hx}}, \mathbf{u})$. Furthermore, given that the Jacobian is lower-triangular because of mutually-independent noise, we can efficiently compute its determinant as the product of diagonal terms. Learning the inverse dynamics won't cause computational issues as well. 
\vspace{-0.5cm}
\paragraph{Factorized Inference} We infer the posteriors of each time step $q(\hat{\mathbf{z}}_t \vert \mathbf{x}_t, \hat{\boldsymbol{\theta}}_k^{\text{obs}})$, using only the observation at that time step and with the change factors $\boldsymbol{\theta}_k^{\text{obs}}$ allocated for global changes. Recurrent inference \citep{yao2021learning} is not used because in Eq.~\ref{eq:model}, $\mathbf{x}_t$ preserves all the information of the current system states so the joint probability $q(\mathbf{\hat z}_{1:T}|\mathbf{x}_{1:T})$ can be factorized into into product of these terms. 

\paragraph{Change Representation}
To allow transfer in a new environment, instead of learning separate transformation or dynamics in each domain \citep{yao2021learning}, we learn to embed domain index $\mathbf{u}_k$ into low-dimensional change factors $(\hat{\boldsymbol{\theta}}_k^{\text{dyn}}, \hat{\boldsymbol{\theta}}_k^{\text{obs}})$ and insert them as inputs to the inverse dynamics function, or the encoder/decoder function, respectively. We make the best use of the multiple domains created by distribution changes, and efficiently correct distribution shits with only a few samples from the target domain, since only the low-dimensional change factors need to be re-estimated.

\begin{algorithm}[htb]
   \caption{\textsc{CorrectShift}$(\hat{f}_i, \hat{\boldsymbol{\theta}}_k, \hat{p}({\epsilon_i}), \hat{g})$}
   \label{alg:correct}
\begin{algorithmic}
   \STATE {\bfseries Input:} Data $\{\mathbf{x}_t\}_{t=1}^T$ for a new environment $k^\ast$. 
   \STATE Initialize change factors $\hat{\boldsymbol{\theta}}_{k^\ast}$ for environment $k^\ast$ and set as model parameters. Freeze other parameters. 
\WHILE{$\hat{\boldsymbol{\theta}}_{k^\ast}$ does not converge}
   \STATE $\hat{\boldsymbol{\theta}_{k^\ast}} \leftarrow \hat{\boldsymbol{\theta}_{k^\ast}} + \alpha \nabla_{\hat{\boldsymbol{\theta}_{k^\ast}}} \mathcal{L}_{\text{ELBO}}(\{\mathbf{x}_t\}_{t=1}^T, \hat{\boldsymbol{\theta}_{k^\ast}})$
 \ENDWHILE
\STATE \textbf{return} the corrected model $(\hat{f}_i, \hat{\boldsymbol{\theta}}_{k^\ast}, \hat{p}({\epsilon_i}), \hat{g})$
 \end{algorithmic}
\end{algorithm}

We train the VAE using the ELBO objective $\mathcal{L}_{\text{ELBO}} = 
\frac{1}{N} \sum_{i \in N} \mathcal{L}_{\text{Recon}} - \beta \mathcal{L}_{\text{KLD}}$, in which we use mean-squared error (MSE) for the reconstruction likelihood $\mathcal{L}_{\text{Recon}}$, and use sampling approach to estimate the $\mathcal{L}_{\text{KLD}}$, since with a learned modular prior, KLD does not have an explicit form.

%
%
%

\section{Experimental Results}
We comparatively evaluate LiLY on a number of time-series datasets. We aim to answer the following questions:
\vspace{-2mm}
\begin{enumerate}[leftmargin=*] \itemsep0em

\item Does LiLY reliably learn temporally-causal latent processes from scratch under different distribution changes? 

\item Is history/nonstationary information necessary for the identifiability of latent causal variables? 

\item How do assumptions in previous nonlinear ICA (i.e.,\ independent sources, nonstationary noises) distort the identifiability results if there are time-delayed causal relations between the latent factors, or if the nonstationary cases involve time-varying causal relations?

\end{enumerate}

\paragraph{Evaluation Metrics}

We report Mean Correlation Coefficient (MCC) on the validation dataset. MCC is a standard metric in the ICA literature for continuous variables which measure the identifiability of the learned latent causal processes. MCC is close to 1 when latent variables are  identifiable up to permutation and componentwise invertible transformation in the noiseless case. To evaluate the transfer performances, we visualize the learned change factors $\boldsymbol{\theta}_k$ with a few samples (16 in our experiments) from the target segment, and compare them with the true change factors. 
 
\paragraph{Baselines}

We used nonlinear ICA baselines: \textbf{(1)} BetaVAE \citep{higgins2016beta} which ignores both history and nonstationarity information;  \textbf{(2)} iVAE \citep{khemakhem2020variational} and TCL \citep{hyvarinen2016unsupervised} which leverage \underline{nonstationarity} to establish identifiability but assumes \underline{independent factors}, and \textbf{(3)} SlowVAE \citep{klindt2020towards} and PCL \citep{hyvarinen2017nonlinear} which exploits \underline{temporal constraints} but assumes \underline{independent sources} and \underline{stationary processes}, and \textbf{(4)} LEAP \citep{yao2021learning} which assumes \underline{nonstationary}, \underline{causal processes} but models all types of distribution changes in \underline{nonstationary noise}.  
\subsection{Synthetic Experiments}
We generate synthetic datasets that satisfy our identifiability conditions in the theorems following the procedures described in Appendix~\ref{ap:synthetic}. In particular, we consider three representative simulation settings respectively to validate the identifiability results under fixed causal dynamics, changing causal dynamics, and modular distribution shift which contains fixed dynamics, changing dynamics and global changes together in the latent processes. For synthetic datasets with fixed and changing causal dynamics, we set latent size $n = 8$. For the modular shift dataset, we add one dimension for global observation changes. The lag number of the process is set to $L = 2$. The mixing function $g$ is a random three-layer MLP with LeakyReLU units.

\subsubsection{Comparisons with Baselines}

As presented in Table~\ref{tab:syn-results}, our framework can recover the latent processes under fixed dynamics (heterogeneous noise model), under changing causal dynamics and under modular distribution shifts. The baselines that do not exploit history or nonstationarity cannot recover the latent processes. SlowVAE and PCL distort the results due to independent source assumptions. LEAP, although exploits both the history and nonstationarity information, considers only limited nonstationary cases in which the noise distributions are changing, thus cannot recover the causal latent processes under changing causal dynamics and modular distribution shifts. 

\vspace{-0.25cm}
\begin{table}[ht]
\centering
\caption{MCC scores and their standard deviations for the three simulation settings over 3 random seeds. Note: The symbol ``--'' represents that this method is not applicable to this dataset.}
\label{tab:syn-results}
\resizebox{0.975\linewidth}{!}{%
\begin{tabular}{@{}cccc@{}}
\toprule
\multicolumn{1}{l}{} & \multicolumn{3}{c}{\textbf{Simulation Setting}} \\ \midrule
 &
  \begin{tabular}[c]{@{}c@{}}Fixed \\ Causal Dynamics\end{tabular} &
  \begin{tabular}[c]{@{}c@{}}Changing \\ Causal Dynamics\end{tabular} &
  \begin{tabular}[c]{@{}c@{}}Modular \\ Shift\end{tabular} \\ \midrule
\underline{\bf LiLY}    &   0.954 \textpm 0.009          &    0.958 \textpm 0.017         &   0.993 \textpm 0.001        \\
LEAP    &   --       &  0.726 \textpm 0.187           &  0.657 \textpm 0.108          \\ 
SlowVAE &  0.411 \textpm 0.022          &  0.511 \textpm 0.062           &  0.406 \textpm 0.045          \\
PCL     & 0.516 \textpm 0.043            &  0.599 \textpm 0.041           &  0.564 \textpm 0.049          \\
i-VAE   &   --       &   0.581 \textpm 0.083          &   0.557 \textpm 0.005         \\
TCL     &   --       &   0.399 \textpm 0.021          &  0.297 \textpm 0.078          \\
$\beta$-VAE &  0.353 \textpm 0.001           &  0.523 \textpm 0.009           &   0.433 \textpm 0.045         \\ \hline
\end{tabular}%
}
\end{table}

\begin{figure*}[ht]
\begin{minipage}{0.3\linewidth}
\centering  
\includegraphics[width=\textwidth]{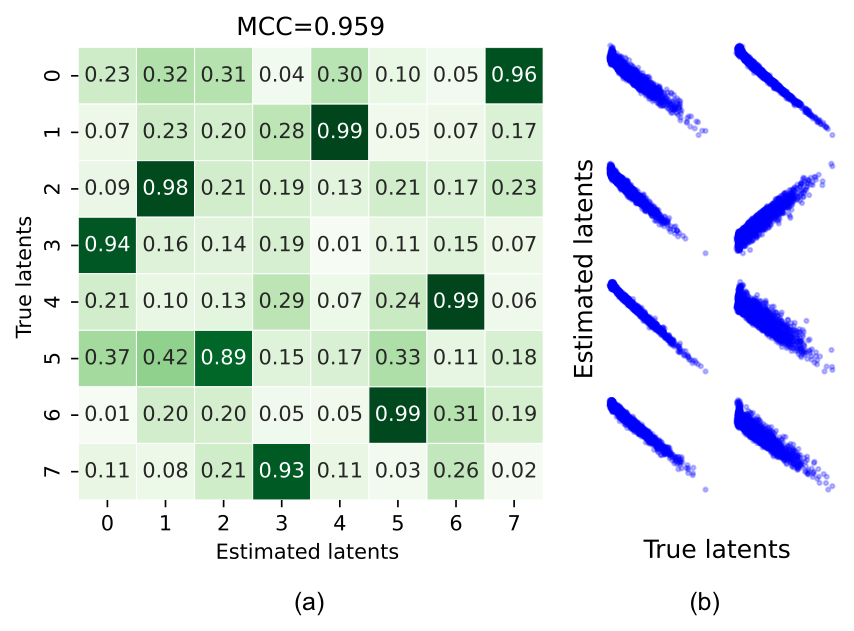}\vspace{-0.5cm}
\caption{Fixed Causal Dynamics.}
\label{fig:fixed}
\end{minipage}%
\hfill
\begin{minipage}{0.3\linewidth}
\includegraphics[width=\textwidth]{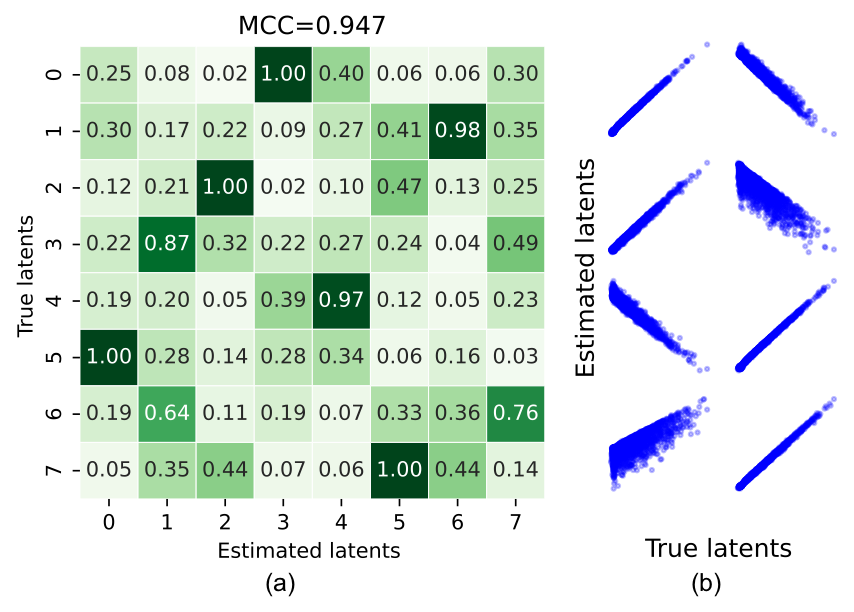}
\vspace{-0.4cm}
\caption{Changing Causal Dynamics.}
\label{fig:change}
\end{minipage}%
\hfill
\begin{minipage}{0.35\linewidth}
\includegraphics[width=\textwidth]{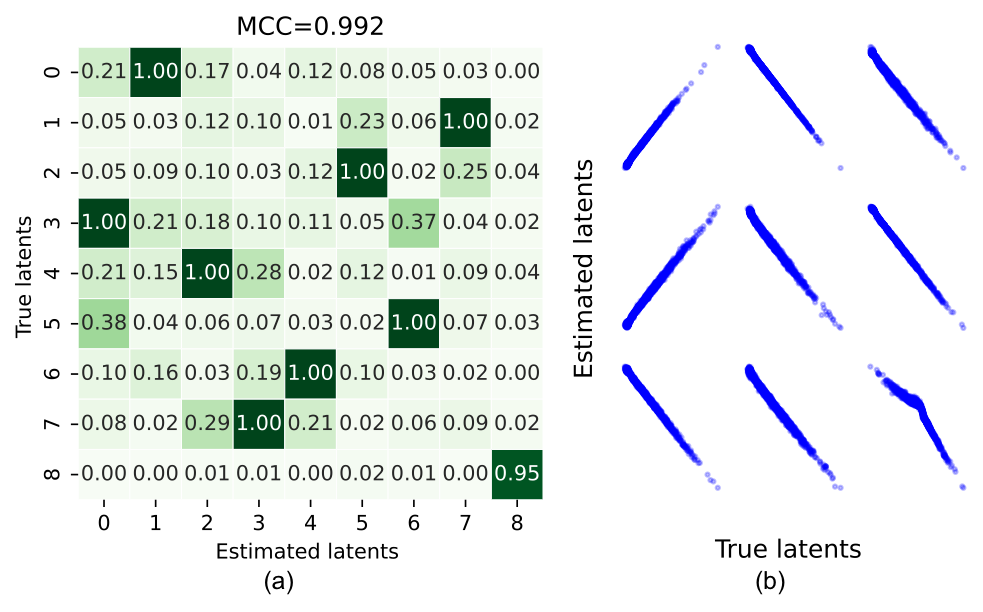}
\vspace{-0.4cm}
\caption{Modular Distribution Shifts.}
\label{fig:modular}
\end{minipage}
\end{figure*}

\subsubsection{Fixed Causal Dynamics}\label{sec:fixed}

We use a heterogeneous noise model in Eq.~\ref{eq:heteo} with standard Gaussian noise terms as the latent processes. The transition function $q_k$ is a random two-layer MLP with LeakyReLU units, and the $b_k$ a mean function of its history $\mathbf{z}_{t-1}$. The latent dynamics is identifiable since the history modulates both the mean and variance of the transition distribution. 
Fig.~\ref{fig:fixed} gives the results on heterogeneous noise datasets. The latent processes are successfully recovered, as indicated by high MCC for the casually-related factors. Panel (b) suggests that the latent causal variables are estimated up to permutation and componentwise invertible transformation. 

\subsubsection{Changing Causal Dynamics}\label{sec:change}

We use a Gaussian additive noise model with changes in the influencing strength as the latent processes. It has been shown in \textbf{Proposition} 1 that this latent dynamics, if without changes, is unidentifiable. We use a random two-layer MLP with LeakyReLU units as the transition function. To add changes, we vary the values of the first layer of the MLP across segments. In Fig.~\ref{fig:change}, the latent processes are successfully recovered, as indicated by high MCC and the recovered latent factors, which shows the further benefits brought by distribution changes. 
\vspace{-0.1cm}
\subsubsection{Modular Distribution Shift}
To partition the latent space, we allocate six latent processes (variables) to fixed heterogeneous noise dynamics as in Sec.~\ref{sec:fixed}, two latent processes following changing causal dynamics in Sec.~\ref{sec:change}, and the last latent variable is sampled from Gaussian distribution whose mean and variance vary by segment index. The results for the modular shift dataset are in Fig.~\ref{fig:modular}. Similarly, the latent processes are recovered, as indicated by (a) high MCC and (b) invertible mappings between the estimated and true latent variables. 
\vspace{-0.1cm}

\subsection{Causal Discovery from Videos}

\begin{figure}[ht]
    \centering
    \includegraphics[width=\linewidth]{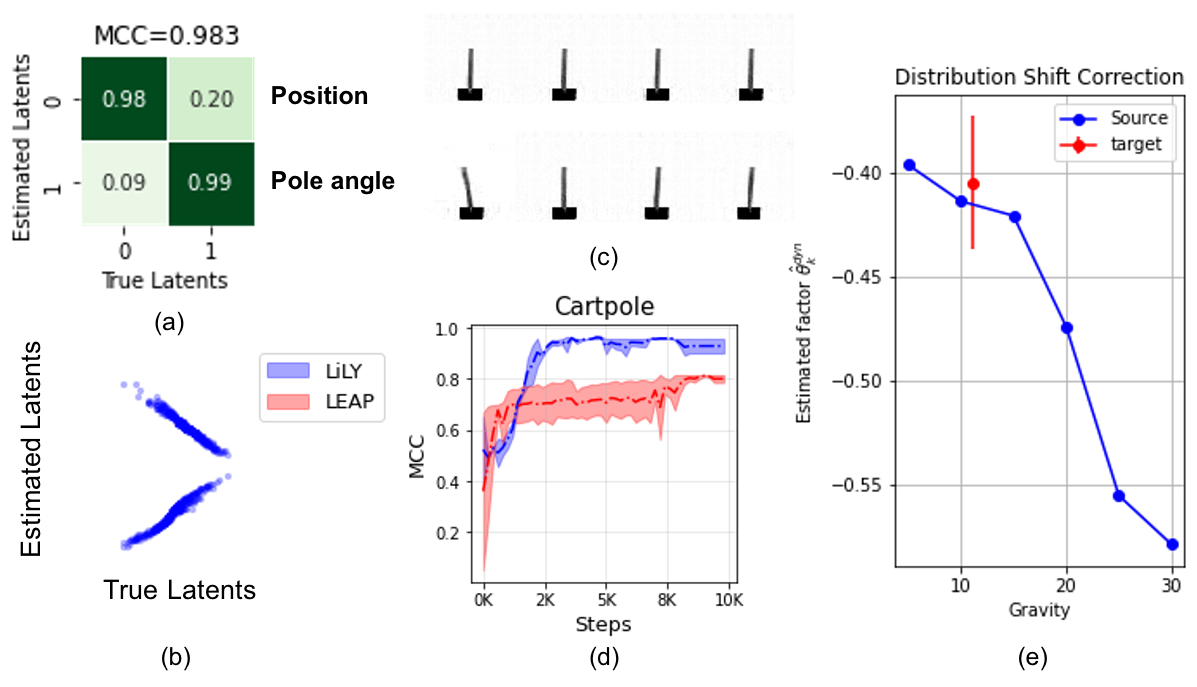}
    \vspace{-0.5cm}
    \caption{Modified Cartpole dataset results: (a) MCC for causally-related factors; (b) scatterplots between estimated and true factors; (c) latent traversal on a fixed video frame; (d) baseline comparisons with LEAP \citep{yao2021learning}, and (e) domain shift correction.  \label{fig:cartpole}}
\end{figure}

We evaluate LiLY on the modified cartpole \citep{huang2021adarl} video dataset and compare the performances with LEAP. Modified Cartpole is a nonlinear dynamical system with cart positions $x_t$ and pole angles $\theta_t$ as the true state variables. The dataset has 6 source domains with different gravity values $g= \{5, 10, 15, 20, 25, 30\}$. Together with the 2 discrete actions (i.e., left and right), we have 12 segments of data with changing causal dynamics. We fit LiLY with two-dimensional change factors $\boldsymbol{\theta}_k^{\text{dyn}}$. We set the latent size $n=8$ and the lag number $L=2$. In Fig.~\ref{fig:cartpole}, the latent causal processes are recovered, as seen from (a) high MCC for the latent causal processes; (b) the latent factors are estimated up to componentwise transformation; (c) the latent traversals confirm that the two latent causal variables correspond to the position and pole angle, and (d) our model outperforms LEAP, which only considers nonstationarity in noise distributions. 

We visualize the learned change factors in panel (e). We also plot the estimated change factor in a new environment ($g = 11.15$) with 16 samples over 3 random seeds. The positive association between the true and estimated change factors indicates that meaningful change representation in the source environment is learned. The mean of the estimated change factor in the target environment is aligned with the true value, showing that we can efficiently adapt with a few samples from the target environment.  

\vspace{-0.1cm}

\section{Conclusion}

We established theories for the identifiability of latent causal dynamics from their observed nonlinear mixtures under fixed dynamics and unknown distribution shifts. The main limitation of this work is that instantaneous causal influences between latent causal processes are not allowed. We acknowledge that there exist situations where the resolution is low and there appears to be instantaneous dependence. However, there are several pieces of work dealing with causal discovery from measured time series in such situations; see. e.g., \cite{granger1987implications, Subsampling_ICML15, Danks14, Aggre_UAI17}. Extending our theories and framework to address the issue of instantaneous dependency or instantaneous causal relations in the latent structure will be one line of our future work.
\bibliography{example_paper}
\bibliographystyle{icml2022}
\clearpage

\onecolumn
\icmltitle{Supplement to: ``Learning Latent Causal Dynamics''}
	
\newcommand{\beginsupplement}{%
	\setcounter{table}{0}
	\renewcommand{\thetable}{S\arabic{table}}%
	\setcounter{figure}{0}
	\renewcommand{\thefigure}{S\arabic{figure}}%
	\setcounter{algorithm}{0}
	\renewcommand{\thealgorithm}{S\arabic{algorithm}}%
	\setcounter{section}{0}
	\renewcommand{\thesection}{S\arabic{section}}%
	\setcounter{theorem}{0}
	\renewcommand{\thetheorem}{S\arabic{theorem}}%
	\setcounter{proposition}{0}
	\renewcommand{\theproposition}{S\arabic{proposition}}%
	\setcounter{corollary}{0}
	\renewcommand{\thecorollary}{S\arabic{corollary}}%
	\renewcommand{\thelemma}{S\arabic{lemma}}%
}

\beginsupplement



\section{Identifiability Theory}\label{ap:theory}

The observed variables were generated according to : 
\begin{equation} \label{Eq:generation}
\mathbf{x}_t = \mathbf{g}(\mathbf{z}_t),    
\end{equation}
in which  $\mathbf{g}$ is invertible, and $z_{it}$, as the $i$th component of  $\mathbf{z}_t$, is generated by (some) components of $\mathbf{z}_{t-1}$ and noise $E_{it}$. $E_{1t}, E_{2t}, ..., E_{nt}$ are mutually independent. In other words, the components of $\mathbf{z}_{t}$ are mutually independent conditional on $\mathbf{z}_{t-1}$. Let $\eta_{kt} \triangleq \log p(z_{kt} | \mathbf{z}_{t-1})$.  Assume that $\eta_k(t)$ is twice differentiable in $z_{kt}$ and is differentiable in $z_{l,t-1}$, $l=1,2,...,n$. Note that the parents of $z_{kt}$ may be only a subset of $\mathbf{z}_{t-1}$; if $z_{l,t-1}$ is not a parent of $z_{kt}$, then $\frac{\partial \eta_k}{\partial z_{l,t-1}} = 0$.

\subsection{Proof for Theorem 1}

\begin{theorem} [Identifiablity under a Fixed Temporal Causal Model] \label{Theo1} 
Suppose there exists invertible function $\mathbf{f}$, which is the estimated mixing function (i.e., we use $\mathbf{f}$ and $\hat{\mathbf{g}}$ interchangeably in Appendix) that maps $\mathbf{x}_t$ to $\hat{\mathbf{z}}_t$, i.e., 
\begin{equation} \label{Eq:invert}
    \hat{\mathbf{z}}_t = \mathbf{f}(\mathbf{x}_t)
\end{equation}
such that the components of $\hat{\mathbf{z}}_t$ are mutually  independent conditional on $\hat{\mathbf{z}}_{t-1}$. 
Let 
\begin{equation} \label{Eq:v}
\mathbf{v}_{kt} \triangleq \Big(\frac{\partial^2 \eta_{kt}}{\partial z_{kt} \partial z_{1,t-1}}, \frac{\partial^2 \eta_{kt}}{\partial z_{kt} \partial z_{2,t-1}}, ..., \frac{\partial^2 \eta_{kt}}{\partial z_{kt} \partial z_{n,t-1}} \Big)^\intercal ,~~~ \mathring{\mathbf{v}}_{kt} \triangleq \Big(\frac{\partial^3 \eta_{kt}}{\partial z_{kt}^2 \partial z_{1,t-1}}, \frac{\partial^3 \eta_{kt}}{\partial z_{kt}^2 \partial z_{2,t-1}}, ..., \frac{\partial^3 \eta_{kt}}{\partial z_{kt}^2 \partial z_{n,t-1}} \Big)^\intercal.    
\end{equation}
If for each value of $\mathbf{z}_t$, $\mathbf{v}_{1t}, \mathring{\mathbf{v}}_{1t}, \mathbf{v}_{2t}, \mathring{\mathbf{v}}_{2t}, ..., \mathbf{v}_{nt}, \mathring{\mathbf{v}}_{nt}$, as 
$2n$ vector functions in $z_{1,t-1}$, $z_{2,t-1}$, ..., $z_{n,t-1}$, are linearly independent, then $\mathbf{z}_{t}$ must be an invertible, component-wise transformation of a permuted version of $\hat{\mathbf{z}}_t$.
\end{theorem}

\begin{proof}
Combining Eq.~\ref{Eq:generation} and Eq.~\ref{Eq:invert} gives $\mathbf{z}_t = \mathbf{g}^{-1}(\mathbf{f}^{-1}(\hat{\mathbf{z}}_t)) = \mathbf{h}(\hat{\mathbf{z}}_t)$, where $\mathbf{h} \triangleq \mathbf{g}^{-1}\circ \mathbf{f}^{-1}$. 
 Since both $\mathbf{f}$ and $\mathbf{g}$ are invertible, $\mathbf{h}$ is invertible.  Let $\mathbf{H}_t$ be the Jacobian matrix of the transformation $h(\hat{\mathbf{z}}_t)$, and denote by $\mathbf{H}_{kit}$ its $(k,i)$th entry.
 
 First, it is straightforward to see that if the components of $\hat{\mathbf{z}}_t$ are mutually independent conditional on $\hat{\mathbf{z}}_{t-1}$, then for any $i\neq j$, $\hat{z}_{it}$ and $\hat{z}_{jt}$ are conditionally independent given $\hat{\mathbf{z}}_{t-1} \cup(\hat{\mathbf{z}}_{t}\setminus \{\hat{z}_{it}, \hat{z}_{jt}\})$. Mutual independence of the components of $\hat{\mathbf{z}}_t$ conditional on $\hat{\mathbf{z}}_{t-1}$ implies that $\hat{z}_{it}$ is independent from $\hat{\mathbf{z}}_{t}\setminus \{\hat{z}_{it}, \hat{z}_{jt}\}$ conditional on $\hat{\mathbf{z}}_{t-1}$, i.e.,
 $$p(\hat{z}_{it} \,|\, \hat{\mathbf{z}}_{t-1}) = p(\hat{z}_{it} \,|\, \hat{\mathbf{z}}_{t-1}\cup(\hat{\mathbf{z}}_{t}\setminus \{\hat{z}_{it}, \hat{z}_{jt}\})).$$
 At the same time, it also implies $\hat{z}_{it}$ is independent from $\hat{\mathbf{z}}_{t}\setminus \{\hat{z}_{it}\}$ conditional on $\hat{\mathbf{z}}_{t-1}$, i.e., 
 $$p(\hat{z}_{it} \,|\, \hat{\mathbf{z}}_{t-1}) = p(\hat{z}_{it} \,|\, \hat{\mathbf{z}}_{t-1}\cup(\hat{\mathbf{z}}_{t}\setminus \{\hat{z}_{it}\})).$$
 Combining the above two equations gives $p(\hat{z}_{it} \,|\, \hat{\mathbf{z}}_{t-1}\cup(\hat{\mathbf{z}}_{t}\setminus \{\hat{z}_{it}\})) = p(\hat{z}_{it} \,|\,\hat{\mathbf{z}}_{t-1} \cup(\hat{\mathbf{z}}_{t}\setminus \{\hat{z}_{it}, \hat{z}_{jt}\}))$, i.e., for $i\neq j$, $\hat{z}_{it}$ and $\hat{z}_{jt}$ are conditionally independent given $\hat{\mathbf{z}}_{t-1} \cup(\hat{\mathbf{z}}_{t}\setminus \{\hat{z}_{it}, \hat{z}_{jt}\})$.
 
 We then make use of the fact that if  $\hat{z}_{it}$ and $\hat{z}_{jt}$ are conditionally independent given $\hat{\mathbf{z}}_{t-1} \cup(\hat{\mathbf{z}}_{t}\setminus \{\hat{z}_{it}, \hat{z}_{jt}\})$, then 
 $$\frac{\partial^2 \log p(\hat{\mathbf{z}}_t, \hat{\mathbf{z}}_{t-1})}{\partial \hat{z}_{it} \partial \hat{z}_{jt}} = 0, $$ 
 assuming the cross second-order derivative exists \cite{Spantini2018InferenceVL}. Since $p(\hat{\mathbf{z}}_t, \hat{\mathbf{z}}_{t-1}) = p(\hat{\mathbf{z}}_t \,|\, \hat{\mathbf{z}}_{t-1})p(\hat{\mathbf{z}}_{t-1})$ while $p(\hat{\mathbf{z}}_{t-1})$ does not involve $\hat{z}_{it}$ or $\hat{z}_{jt}$, the above equality is equivalent to 
 \begin{equation} \label{Eq:iszero}
     \frac{\partial^2 \log p(\hat{\mathbf{z}}_t \,|\,\hat{\mathbf{z}}_{t-1})}{\partial \hat{z}_{it} \partial \hat{z}_{jt}} = 0.
 \end{equation}
 
 The Jacobian matrix of the mapping from $(\mathbf{x}_{t-1}, \hat{\mathbf{z}}_t)$ to $(\mathbf{x}_{t-1}, \mathbf{z}_t)$ is $\begin{bmatrix}\mathbf{I} & \mathbf{0} \\ * & \mathbf{H}_t \end{bmatrix}$, where $*$ stands for a matrix, and the (absolute value of the) determinant of this Jacobian matrix is $|\mathbf{H}_t|$. Therefore $p(\hat{\mathbf{z}}_t, \mathbf{x}_{t-1}) = p({\mathbf{z}}_t, \mathbf{x}_{t-1})\cdot |\mathbf{H}_t|$. Dividing both sides of this equation by $p(\mathbf{x}_{t-1})$ gives 
 \begin{equation} \label{Eq:J_trans}
 p(\hat{\mathbf{z}}_t \,|\, \mathbf{x}_{t-1}) = p({\mathbf{z}}_t \,|\, \mathbf{x}_{t-1}) \cdot |\mathbf{H}_t|. 
 \end{equation}
 Since $p({\mathbf{z}}_t \,|\, {\mathbf{z}}_{t-1}) = p({\mathbf{z}}_t \,|\, \mathbf{g}({\mathbf{z}}_{t-1})) = p({\mathbf{z}}_t \,|\, {\mathbf{x}}_{t-1})$ and similarly   $p(\hat{\mathbf{z}}_t \,|\, \hat{\mathbf{z}}_{t-1}) = p(\hat{\mathbf{z}}_t \,|\, {\mathbf{x}}_{t-1})$, Eq.~\ref{Eq:J_trans} tells us
 \begin{equation}
     \log p(\hat{\mathbf{z}}_t \,|\, \hat{\mathbf{z}}_{t-1}) = \log p({\mathbf{z}}_t \,|\, {\mathbf{z}}_{t-1}) + \log |\mathbf{H}_t| = \sum_{k=1}^n \eta_{kt} + \log |\mathbf{H}_t|.
 \end{equation}
 Its partial derivative w.r.t. $\hat{z}_{it}$ is 
 \begin{flalign} \nonumber
  \frac{\partial \log p(\hat{\mathbf{z}}_t \,|\, \hat{\mathbf{z}}_{t-1})}{\partial \hat{z}_{it}} &=  \sum_{k=1}^n \frac{\partial \eta_{kt} }{\partial z_{kt}} \cdot \frac{\partial z_{kt}}{\partial \hat{z}_{it}} - \frac{\partial \log |\mathbf{H}_t|}{\partial \hat{z}_{it}} \\ \nonumber
  &= \sum_{k=1}^n \frac{\partial \eta_{kt}}{\partial z_{kt}} \cdot \mathbf{H}_{kit} - \frac{\partial \log |\mathbf{H}_t|}{\partial \hat{z}_{it}}.
 \end{flalign}
  Its second-order cross derivative is
 \begin{flalign} \label{Eq:cross}
  \frac{\partial^2 \log p(\hat{\mathbf{z}}_t \,|\, \hat{\mathbf{z}}_{t-1})}{\partial \hat{z}_{it} \partial \hat{z}_{jt}}
  &= \sum_{k=1}^n \Big( \frac{\partial^2 \eta_{kt}}{\partial z_{kt}^2 } \cdot \mathbf{H}_{kit}\mathbf{H}_{kjt} + \frac{\partial \eta_{kt}}{\partial z_{kt}} \cdot \frac{\partial \mathbf{H}_{kit}}{\partial \hat{z}_{jt}} \Big)- \frac{\partial^2 \log |\mathbf{H}_t|}{\partial \hat{z}_{it} \partial \hat{z}_{jt}}.
 \end{flalign}
 
 The above quantity is always 0 according to Eq.~\ref{Eq:iszero}. Therefore, for each $l=1,2,...,n$ and each value $z_{l,t-1}$,  its partial derivative w.r.t.
 $z_{l,t-1}$ is always 0. That is,
 \begin{flalign}
  \frac{\partial^3 \log p(\hat{\mathbf{z}}_t \,|\, \hat{\mathbf{z}}_{t-1})}{\partial \hat{z}_{it} \partial \hat{z}_{jt} \partial z_{l,t-1}}
  &= \sum_{k=1}^n \Big( \frac{\partial^3 \eta_{kt}}{\partial z_{kt}^2 \partial z_{l,t-1}} \cdot \mathbf{H}_{kit}\mathbf{H}_{kjt} + \frac{ \partial^2 \eta_{kt}}{\partial z_{kt} \partial z_{l,t-1}}  \cdot \frac{\partial \mathbf{H}_{kit}}{\partial \hat{z}_{jt} } \Big) \equiv 0,
 \end{flalign}
 where we have made use of the fact that entries of $\mathbf{H}_t$ do not depend on $z_{l,t-1}$. 
 
 If for any value of $\mathbf{z}_t$, $\mathbf{v}_{1t}, \mathring{\mathbf{v}}_{1t}, \mathbf{v}_{2t}, \mathring{\mathbf{v}}_{2t}, ..., \mathbf{v}_{nt}, \mathring{\mathbf{v}}_{nt}$ are linearly independent, to make the above equation hold true, one has to set $\mathbf{H}_{kit}\mathbf{H}_{kjt} = 0$ or $i\neq j$. That is, in each row of $\mathbf{H}_t$ there is only one non-zero entry. Since $h$ is invertible, then $\mathbf{z}_{t}$ must be an invertible, component-wise transformation of a permuted version of $\hat{\mathbf{z}}_t$.
\end{proof}

The linear independence condition in Theorem \ref{Theo1} is the core condition to guarantee the identifiability of $\mathbf{z}_t$ from the observed $\mathbf{x}_t$. Roughly speaking, for a randomly chosen conditional density function $p(z_{kt}\,|\,\mathbf{z}_{t-1})$, the chance for this constraint to hold on its second- and third-order partial derivatives is slim. For illustrative purposes, below we make this claim more precise, by considering a specific unidentifiable case, in which the noise terms in $\mathbf{z}_t$ are additive Gaussian, and two identifiable cases, in which $\mathbf{z}_t$ has additive, heterogeneous noise or follows some linear, non-Gaussian temporal process. 

Let us start with an unidentifiable case. 
If all $z_{kt}$ follow the additive noise model with Gaussian noise terms, i.e., 
\begin{equation} \label{Eq:Gaussian_case}
    \mathbf{z}_t = \mathbf{q}(\mathbf{z}_{t-1}) + \mathbf{E}_t,
\end{equation}
where $\mathbf{q}$ is a transformation and the components of the Gaussian vector $\mathbf{E}_t$ are independent and also independent from $\mathbf{z}_{t-1}$. Then  $\frac{\partial^2 \eta_{kt}}{\partial z_{kt}^2}$ is constant, and $\frac{\partial^3 \eta_{kt}}{\partial z_{kt}^2 \partial z_{l,t-1}} \equiv 0$, violating the linear independence condition. In the following proposition we give some alternative solutions and verify the unidentifiability in this case.

\begin{proposition} [Unidentifiability under Gaussian noise] 
Suppose $\mathbf{x}_t$ was generated according to Eq.~\ref{Eq:generation} and Eq.~\ref{Eq:Gaussian_case},
where the components of $\mathbf{E}_t$ are mutually independent Gaussian and also independent from $\mathbf{z}_{t-1}$. Then any $\hat{\mathbf{z}}_t = \mathbf{D}_1 \mathbf{U} \mathbf{D}_2 \cdot {\mathbf{z}}_t$, where $\mathbf{D}_1$ is an arbitrary non-singular diagonal matrix, $\mathbf{U}$ is an arbitrary orthogonal matrix, and $\mathbf{D}_2$ is a diagonal matrix with $\mathbb{V}ar^{-1/2}(E_{kt})$ as its $k$th  diagonal entry, is a valid solution to satisfy the condition that the components of $\hat{\mathbf{z}}_t$ are mutually independent conditional on $\hat{\mathbf{z}}_{t-1}$.
\end{proposition}
\begin{proof}
In this case we have 
$$ \hat{\mathbf{z}}_t = \mathbf{D}_1 \mathbf{U} \mathbf{D}_2 \cdot \mathbf{q}(\mathbf{z}_{t-1}) + \mathbf{D}_1 \mathbf{U} \mathbf{D}_2\cdot \mathbf{E}_t.$$
It is easy to verify that the components of $\mathbf{D}_1 \mathbf{U} \mathbf{D}_2\cdot \mathbf{E}_t$ are mutually independent and are independent from $\mathbf{D}_1 \mathbf{U} \mathbf{D}_2 \cdot \mathbf{q}(\mathbf{z}_{t-1})$. As a consequence, $\hat{\mathbf{z}}_t$ are mutually independent conditional on $\hat{\mathbf{z}}_{t-1}$.
\end{proof}

Now let us consider some cases in which the latent temporally processes $\mathbf{z}_t$ are naturally identifiable under some technical conditions. Let us first consider the case where $z_{kt}$ follows a heterogeneous noise process, in which the noise variance depends on its parents:
\begin{equation} \label{Eq:heteo}
    a_{kt} = q_k(\mathbf{z}_{t-1}) + \frac{1}{b_k(\mathbf{z}_{t-1})}E_{kt}.
\end{equation}
Here we assume $E_{kt}$ is standard Gaussian and $E_{1t}, E_{2t}, .., E_{nt}$ are mutually independent and independent from $\mathbf{z}_{t-1}$. $\frac{1}{b_k}$, which depends on $\mathbf{z}_{t-1}$, is the standard deviation of the noise in $z_{kt}$. (For conciseness, we drop the argument of $b_k$ and $q_k$ when there is no confusion.) Note that in this model, if $q_k$ is 0 for all $k=1,2,...,n$, it reduces to a multiplicative noise model. 
The identifiability result of $\mathbf{z}_t$ is established in the following proposition.

\begin{corollary} [Identifiablity under Heterogeneous Noise] 
Suppose $\mathbf{x}_t$ was generated according to Eq.~\ref{Eq:generation} and Eq.~\ref{Eq:heteo}. Suppose Eq.~\ref{Eq:invert} holds true. If  $b_k\cdot \frac{\partial b_k}{\partial \mathbf{z}_{t-1}}$ and $b_k\cdot \frac{\partial b_k}{\partial \mathbf{z}_{t-1}} (z_{kt} - q_{k}) - b_k^2\cdot \frac{\partial q_{k}}{\partial \mathbf{z}_{t-1}}$, with $k=1,2,...,n$, which are in total $2n$ function vectors in $\mathbf{z}_{t-1}$, 
are linearly independent, then $\mathbf{z}_{t}$ must be an invertible, component-wise transformation of a permuted version of $\hat{\mathbf{z}}_t$.

\end{corollary}

\begin{proof}
Under the assumptions, one can see that 
$$ \eta_{kt} = \log p(z_{kt}\,|\, \mathbf{z}_{t-1}) = -\frac{1}{2}\log(2\pi) + \log b_k - \frac{b_k^2}{2} (z_{kt} - q_{k} )^2.$$
Consequently, one can find
\begin{flalign}
 \nonumber
\frac{\partial^3 \eta_{kt}}{\partial z_{kt}^2 \partial z_{l,t-1}} &= -b_k\cdot \frac{\partial b_k}{\partial z_{l,t-1}}, \\ \nonumber
\frac{\partial^2 \eta_{kt}}{\partial z_{kt} \partial z_{l,t-1}} &= -b_k\cdot \frac{\partial b_k}{\partial z_{l,t-1}} (z_{kt} - q_{k}) + b_k^2\cdot \frac{\partial q_{k}}{\partial z_{l,t-1}}.
\end{flalign} 
Then the linear independence of $\mathbf{v}_{kt}$ and $\mathring{\mathbf{v}}_{kt}$ (defined in Eq.~\ref{Eq:v}), with $k=1.,2,...,n$, reduces to the linear independence condition in this proposition. Theorem \ref{Theo1}
 then implies that $\mathbf{z}_{t}$ must be an invertible, component-wise transformation of a permuted version of $\hat{\mathbf{z}}_t$.
\end{proof}

Let us then consider another special case, with linear, non-Gaussian temporal model for $\mathbf{z}_t$: the latent processes follow Eq.~\ref{Eq:Gaussian_case}, with $\mathbf{q}$ being a linear transformation and $E_{kt}$ following a particular class of non-Gaussian distributions. 
The following corollary shows that $\mathbf{z}_t$ is identifiable as long as each $z_{kt}$ receives causal influences from some components of $\mathbf{z}_{t-1}$.
\begin{corollary} [Identifiablity under a Specific Linear, Non-Gaussian Model for Latent Processes]  
Suppose $\mathbf{x}_t$ was generated according to Eq.~\ref{Eq:generation} and Eq.~\ref{Eq:Gaussian_case}, in which $\mathbf{q}$ is a linear transformation and for each $z_{kt}$, there exists at least one $k'$ such that $c_{kk'} \triangleq \frac{\partial z_{kt}}{\partial z_{k',t-1}} \neq 0$. Assume the noise term $E_{kt}$ follows a zero-mean generalized normal distribution:
\begin{equation}
    p(E_{kt}) \propto e^{-\lambda |e_{kt}|^\beta}\textrm{,~~with positive $\lambda$ and }\beta > 2 \textrm{ and } \beta \neq 3.
\end{equation}
Suppose Eq.~\ref{Eq:invert} holds true. Then $\mathbf{z}_{t}$ must be an invertible, component-wise transformation of a permuted version of $\hat{\mathbf{z}}_t$.
\end{corollary}

\begin{proof}
In this case, we have
\begin{flalign}
\frac{\partial^3 \eta_{kt}}{\partial z_{kt}^2 \partial z_{k',t-1}} &= - \lambda \cdot \textrm{sgn}(e_{kt}) \cdot \alpha(\beta-1)(\beta-2)|e_{kt}|^{\beta - 3}c_{kk'}, \\ \nonumber
\frac{\partial^2 \eta_{kt}}{\partial z_{kt} \partial z_{k',t-1}} &= -\lambda  \beta(\beta-1)|e_{kt}|^{\beta - 2}c_{kk'}.
\end{flalign}
We know that $|e_{lt}|^{\beta - 2}$ and $|e_{lt}|^{\beta - 3}$ are linearly independent (because their ratio, $|e_{lt}|$, is not constant). Furthermore, $|e_{lt}|^{\beta - 2}$ and $|e_{lt}|^{\beta - 3}$, with $l=1,2,...,n$, are $2n$ linearly independent functions (because of the different arguments involved).

Suppose there exist $\alpha_{l1}$ and $\alpha_{l2}$, with $l=1,2,...,n$, such that 
\begin{equation} \label{Eq:linear_dep}
\sum_{l=1}^n \big( \alpha_{l1} \mathbf{v}_{lt} + \alpha_{l2} \mathring{\mathbf{v}}_{lt} \big) = 0.
\end{equation}
It is assumed that for each $k=1, 2,..., n$, there exists at least one $k'$ such that $c_{kk'} \neq 0$. Eq.~\ref{Eq:linear_dep} then implies that for any $k$ we have
\begin{equation}
     \alpha_{k1}c_{kk'} |e_{kt}|^{\beta - 2} + \alpha_{k2} c_{kk'}|e_{kt}|^{\beta - 3} + \sum_{l \neq k} \big( \alpha_{l1}c_{lk'} |e_{lt}|^{\beta - 2} + \alpha_{l2} c_{lk'}|e_{lt}|^{\beta - 3} \big) = 0.
\end{equation}
Since $|e_{lt}|^{\beta - 2}$ and $|e_{lt}|^{\beta - 3}$, with $l=1,2,...,n$, are linearly independent and $c_{kk'} \neq 0$, to make the above equation hold, one has to set $\alpha_{k1} = \alpha_{k2} = 0$. As this applies to any $k$, we know that for Eq.~\ref{Eq:linear_dep} to be satisfied, $\alpha_{l1}$ and $\alpha_{l2}$ must be 0, for all $l=1,2,...,n$. That is,  $\mathbf{v}_{1t}, \mathring{\mathbf{v}}_{1t}, \mathbf{v}_{2t}, \mathring{\mathbf{v}}_{2t}, ..., \mathbf{v}_{nt}, \mathring{\mathbf{v}}_{nt}$ are linearly independent. The linear independence condition in Theorem \ref{Theo1} is satisfied.  Therefore 
$\mathbf{z}_{t}$ must be an invertible, component-wise transformation of a permuted version of $\hat{\mathbf{z}}_t$.
\end{proof}



\subsection{Proof for Theorem 2 and 3}

Let $\mathbf{v}_{kt}(u_r)$ be $\mathbf{v}_{kt}$, which is defined in Eq.~\ref{Eq:v}, in the $u_r$ context. Similarly, Let $\mathring{\mathbf{v}}_{kt}(u_r)$ be $\mathring{\mathbf{v}}_{kt}$ in the $u_r$ context. Let 
$$\mathbf{s}_{kt} \triangleq \Big( \mathbf{v}_{kt}(u_1)^\intercal, ..., 
\mathbf{v}_{kt}(u_m)^\intercal, 
\frac{\partial^2 \eta_{kt}({u}_{2})}{\partial z_{kt}^2 } - 
 \frac{\partial^2 \eta_{kt}({u}_{1})}{\partial z_{kt}^2 }, ...,
 \frac{\partial^2 \eta_{kt}({u}_{m})}{\partial z_{kt}^2 } - 
 \frac{\partial^2 \eta_{kt}({u}_{m-1})}{\partial z_{kt}^2 }
 \Big)^\intercal,$$
 $$\mathring{\mathbf{s}}_{kt} \triangleq \Big( \mathring{\mathbf{v}}_{kt}(u_1)^\intercal, ..., 
\mathring{\mathbf{v}}_{kt}(u_m)^\intercal, 
\frac{\partial \eta_{kt}({u}_{2})}{\partial z_{kt} } - 
 \frac{\partial \eta_{kt}({u}_{1})}{\partial z_{kt} }, ...,
 \frac{\partial \eta_{kt}({u}_{m})}{\partial z_{kt} } - 
 \frac{\partial \eta_{kt}({u}_{m-1})}{\partial z_{kt} }
 \Big)^\intercal.$$
As provided below, in our case, the identifiablity of $\mathbf{z}_t$ is guaranteed by the linear independence of the whole function vectors $\mathbf{s}_{kt}$ and $\mathring{\mathbf{s}}_{kt}$, with $k=1,2,...,n$.
However, the identifiability result in Yao et al. (2021) relies on the linear independence of only the last $m-1$ components of $\mathbf{s}_{kt}$ and $\mathring{\mathbf{s}}_{kt}$ with $k=1,2,...,n$; this linear independence is generally a much stronger condition.  

\begin{theorem}[Identifiability under Changing Causal Dynamics]
Suppose the observed processes $\mathbf{x}_t$ was generated by Eq.~\ref{Eq:generation} and that the conditional distribution $p(z_{kt} \,|\, \mathbf{z}_{t-1})$ may change across $m$ values of the context variable $\mathbf{u}$, denoted by $u_1$, $u_2$, ..., $u_m$. Suppose the components of $\mathbf{z}_t$ are mutually independent conditional on $\mathbf{z}_{-1}$ in each context. Assume that the components of $\hat{\mathbf{z}}_t$ produced by Eq.~\ref{Eq:invert} are also mutually independent conditional on $\hat{\mathbf{z}}_{t-1}$. 
If the $2n$ function vectors $\mathbf{s}_{kt}$ and $\mathring{\mathbf{s}}_{kt}$, with $k=1,2,...,n$, are linearly independent, then $\hat{\mathbf{z}}_t$ is a permuted invertible component-wise transformation of $\mathbf{z}_t$. 
\end{theorem}

\begin{proof}
As in the proof of Theorem \ref{Theo1}, because the components of $\hat{\mathbf{z}}_t$ are mutually independent conditional on $\hat{\mathbf{z}}_{t-1}$, we know that 
for $i\neq j$, 
\begin{equation}
 \label{Eq:cross2}
  \frac{\partial^2 \log p(\hat{\mathbf{z}}_t \,|\, \hat{\mathbf{z}}_{t-1}; \mathbf{u})}{\partial \hat{z}_{it} \partial \hat{z}_{jt}}
  = \sum_{k=1}^n \Big( \frac{\partial^2 \eta_{kt}(\mathbf{u})}{\partial z_{kt}^2 } \cdot \mathbf{H}_{kit}\mathbf{H}_{kjt} + \frac{\partial \eta_{kt}(\mathbf{u})}{\partial z_{kt}} \cdot \frac{\partial \mathbf{H}_{kit}}{\partial \hat{z}_{jt}} \Big)- \frac{\partial^2 \log |\mathbf{H}_t|}{\partial \hat{z}_{it} \partial \hat{z}_{jt}} \equiv 0.
\end{equation}
Compared to Eq.~\ref{Eq:cross}, here we allow $p(\hat{\mathbf{z}}_t \,|\, \hat{\mathbf{z}}_{t-1})$ to depend on $\mathbf{u}$.  Since the above equations is always 0, taking its partial derivative w.r.t. $z_{l,t-1}$ gives
 \begin{flalign} \label{Eq:linear_part1}
  \frac{\partial^3 \log p(\hat{\mathbf{z}}_t \,|\, \hat{\mathbf{z}}_{t-1}; \mathbf{u})}{\partial \hat{z}_{it} \partial \hat{z}_{jt} \partial z_{l,t-1}}
  &= \sum_{k=1}^n \Big( \frac{\partial^3 \eta_{kt}(\mathbf{u})}{\partial z_{kt}^2 \partial z_{l,t-1}} \cdot \mathbf{H}_{kit}\mathbf{H}_{kjt} + \frac{ \partial^2 \eta_{kt}( \mathbf{u})}{\partial z_{kt} \partial z_{l,t-1}}  \cdot \frac{\partial \mathbf{H}_{kit}}{\partial \hat{z}_{jt} } \Big) \equiv 0.
 \end{flalign}
Similarly, Using different values for $\mathbf{u}$ in Eq.~\ref{Eq:cross2} take the difference of this equation across them gives
\begin{flalign}
 \nonumber 
  &\frac{\partial^2 \log p(\hat{\mathbf{z}}_t \,|\, \hat{\mathbf{z}}_{t-1}; {u}_{r+1})}{\partial \hat{z}_{it} \partial \hat{z}_{jt}} - \frac{\partial^2 \log p(\hat{\mathbf{z}}_t \,|\, \hat{\mathbf{z}}_{t-1}; {u}_{r+1})}{\partial \hat{z}_{it} \partial \hat{z}_{jt}}
  \\ \label{Eq:cross3} =& \sum_{k=1}^n \Big[ \Big(\frac{\partial^2 \eta_{kt}({u}_{r+1})}{\partial z_{kt}^2 } - 
 \frac{\partial^2 \eta_{kt}({u}_{r})}{\partial z_{kt}^2 } \Big) \cdot \mathbf{H}_{kit}\mathbf{H}_{kjt} + \Big(\frac{\partial \eta_{kt}(u_{r+1})}{\partial z_{kt}} - \frac{\partial \eta_{kt}(u_{r})}{\partial z_{kt}}\Big) \cdot \frac{\partial \mathbf{H}_{kit}}{\partial \hat{z}_{jt}} \Big] \equiv 0.
\end{flalign}

Therefore, if $\mathbf{s}_{kt}$ and $\mathring{\mathbf{s}}_{kt}$, for $k=1,2,...,n$, are linearly independent, $\mathbf{H}_{kit}\mathbf{H}_{kjt}$ has to be zero for all $k$ and $i\neq j$. Then as shown in the proof of Theorem \ref{Theo1}, $\hat{\mathbf{z}}_t$ must be a permuted component-wise invertible transformation of $\mathbf{z}_t$. 
\end{proof}

\begin{theorem}[Identifiability under Observation Changes]
Suppose $\mathbf{x}_t = \mathbf{g}(\mathbf{z}_t)$ and that the conditional distribution $p(z_{k,t} \,|\, \mathbf{u})$ may change across $m$ values of the context variable $\mathbf{u}$, denoted by $u_1$, $u_2$, ..., $u_m$. Suppose the components of $\mathbf{z}_t$ are mutually independent conditional on $\mathbf{u}$ in each context. Assume that the components of $\hat{\mathbf{z}}_t$ produced by Eq.~\ref{eq:invert} are also mutually independent conditional on $\hat{\mathbf{z}}_{t-1}$. 
If the $2n$ function vectors $\mathbf{s}_{k,t}$ and $\mathring{\mathbf{s}}_{k,t}$, with $k=1,2,...,n$, are linearly independent, then $\hat{\mathbf{z}}_t$ is a permuted invertible component-wise transformation of $\mathbf{z}_t$.

\end{theorem}

\begin{proof}
As in the proof of Theorem S2, because $\mathbf{z}_t$ is not dependent on the history $\mathbf{z}_{t-1}$ so are the components of $\hat{\mathbf{z}}_t$, the conditioning on $\hat{\mathbf{z}}_t$ in Eq.~\ref{Eq:cross2} and the following equations can be removed because of the independence. This directly leads to the same conclusion as in Theorem S2. 

\end{proof}

\begin{corollary}[Identifiability under Modular Distribution Shifts] Assume the data generating process in Eq.~\ref{eq:model}. If the three partitioned latent components $\mathbf{z}_t = (\mathbf{z}_t^{\text{fix}}, \mathbf{z}_t^{\text{chg}}, \mathbf{z}_t^{\text{obs}})$ respectively satisfy the conditions in \textbf{Theorem} 1, \textbf{Theorem} 2, and \textbf{Theorem} 3, then $\mathbf{z}_{t}$ must be an invertible, component-wise transformation of a permuted version of $\hat{\mathbf{z}}_t$. 
 
\end{corollary}

\begin{proof}
Because the three partitioned subspaces $(\mathbf{z}_t^{\text{fix}}, \mathbf{z}_t^{\text{chg}}, \mathbf{z}_t^{\text{obs}})$ are conditional independent given the history and domain index, it is straightforward to factorize the joint conditional log density into three components. By using the proof in Theorem 1, 2, and 3, we can directly derive the same quantity as in Eq. 15 or Eq. 24. Therefore, if $\mathbf{s}_{kt}$ and $\mathring{\mathbf{s}}_{kt}$, for $k=1,2,...,n$, are linearly independent, $\mathbf{H}_{kit}\mathbf{H}_{kjt}$ has to be zero for all $k$ and $i\neq j$. Then as shown in the proof of Theorem \ref{Theo1}, $\hat{\mathbf{z}}_t$ must be a permuted component-wise invertible transformation of $\mathbf{z}_t$. 

\end{proof}
\subsection{Comparisons with Existing Identifiablility Theories}
\label{ap:compar}
 The closest work to ours includes (1) \underline{LEAP}~\citep{yao2021learning}, which leverage the changes in noise distribution to disentangle the nonparametric causal processes, (2) \underline{PCL}~\citep{hyvarinen2017nonlinear}, which exploited temporal constraints to separate independent sources, and (3) \underline{SlowVAE}~\citep{klindt2020towards}, which leveraged sparse transition of adjacent video frames to separate independent sources.. 

\paragraph{LEAP} The sources $z_{it}$ in LEAP can have nonparametric, time-delayed causal relations in between. However, LEAP only considers a special case of nonstationarity caused by changes in noise distributions across domains, while our work can allow the causal influencing strength to change across segments. Furthermore, because LEAP assumes all sources' conditional distributions are changed by nonstationary noise, it doesn't exploit the fixed dynamics parts for identifiability. On the other hand, since our work exploits both the distribution changes from the fixed causal dynamics and changing dynamics (also from observation changes), our derived identifiability conditions are generally weaker than \citep{yao2021learning}.

\begin{equation}\label{eq:np-gen}
   \underbrace{ \mathbf{x}_t = g(\mathbf{z}_t) }_{\text{Nonlinear mixing}}, \quad \underbrace{z_{it} = f_i\left(\{z_{j, t-\tau} \vert z_{j, t-\tau} \in \mathbf{Pa}(z_{it}) \}, \epsilon_{it}  \right)}_{\text{Nonparametric transition}} \; with \underbrace{\epsilon_{it} \sim p_{\epsilon_i \vert \mathbf{u}}}_{\text{Nonstationary noise}}.
\end{equation}

\paragraph{PCL}
 The sources $z_{it}$ in PCL were assumed to be mutually independent (see Assumption 1 of Theorem 1 in PCL). In contrast, our identifiability conditions under fixed causal dynamics allow the sources to have time-delayed causal relations in between, which is much more realistic in real-world applications. The underlying processes of PCL are described by \Eqref{eq:comp0}:
\begin{equation}
\small
\label{eq:comp0}
\log p(z_{i,t}|z_{i,t-1})=G(z_{i,t}-\rho z_{i,t-1}) \quad {\rm or} \quad
\log p(z_{i,t}|z_{i,t-1})=-\lambda \left(z_{i,t} - r(z_{i,t-1})\right)^2+{\rm const}.
\end{equation}
where $G$ is some non-quadratic function corresponding to
the log-pdf of innovations, $\rho<1$ is regression coefficient, $r$ is some nonlinear, strictly monotonic regression, and $\lambda$ is a positive precision parameter.

\paragraph{SlowVAE}
 Inspired by slow feature analysis, SlowVAE assumes the underlying sources to have identity transitions with generalized Laplacian innovations described in \Eqref{eq:comp1}:
\begin{equation}
\label{eq:comp1}
 p(\mathbf{z}_t|\mathbf{z}_{t-1})=\prod \limits_{i=1}^d\frac{\alpha \lambda}{2 \Gamma (1/\alpha)}\exp{-(\lambda|z_{i,t}-z_{i,t-1}|^{\alpha})} \quad with \quad \alpha < 2.
\end{equation}
 Our Corollay 2 completes to the Laplacian innovation model above by allowing time-delayed vector autoregressive transitions in the latent process with multiple time lags, and with generalized Gaussian noises. Consequently, temporally causally-related latent processes with linear transition dynamics can thus be modeled and recovered from their nonlinear mixtures with our condition.

\section{Experiment Settings}

\subsection{Datasets}
\subsubsection{Synthetic Dataset Generation}\label{ap:synthetic}
To evaluate the identifiability of our method under different conditions, we generate the synthetic data with 1) fixed causal dynamics; 2) changing causal dynamics and 3) the modular distribution shift. 

\paragraph{Fixed Causal Dynamics}

For the fixed causal dynamics. We generate 100,000 data points according to Eq. \eqref{eq:heteo}, where the latent size is $n=8$, lag number of the process is $L = 2$.
We apply a 2-layer MLP with LeakyReLU as the state transition function. The process noise are sampled from i.i.d. Gaussian distribution ($\sigma=0.1$). The process noise terms are coupled with the history information through multiplication with the average value of all the time-lagged latent variables. 

\paragraph{Changing Causal Dynamics}

We use a Gaussian additive noise model with changes in the influencing strength as the latent processes. To add changes, we vary the values of the first layer of the MLP across the 20 segments and generate 7,500 samples for each segment. The entries of the kernel matrix of the first layer are uniformly distributed between $[-1,1]$ in each domain. 

\paragraph{Modular Distribution Shifts}

The latent space of this dataset is partitioned into 6 fixed dynamics components under the heterogeneous noise model, 2 changing components with changing causal dynamics and 1 component modulated by domain index only. The fixed and changing dynamics components follow the same generating procedures above. The global change component is sampled from i.i.d Gaussian distribution whose mean and variance are modulated by domain index. In particular, distribution mean terms are uniformly sampled between $[-1,1]$ and variance terms are uniformly sampled between $[0.01, 1]$. 

\subsubsection{Real-world Dataset}\label{ap:real}
\paragraph{Modified Cartpole} The Cartpole problem \citep{huang2021adarl} ``consists of a cart and a vertical pendulum attached to the cart using a passive pivot joint. The cart can move left or right. The task is to prevent the vertical pendulum from falling by putting a force on the cart to move it left or right. The action space consists of two actions: moving left or right.'' 

\begin{figure}[h]
 \centering
 \includegraphics[width=0.75\textwidth]{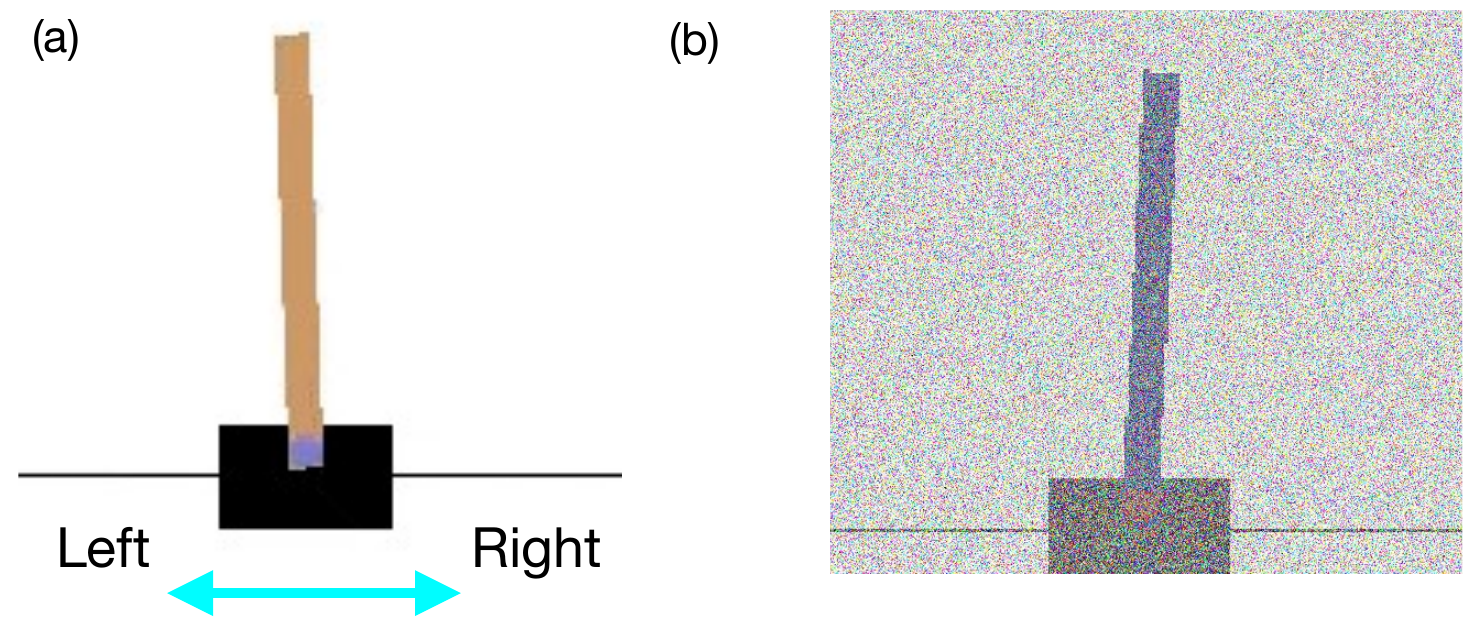}
 \caption{Visual examples of Cartpole game and change factors. (a) Cartpole game; (b) Modified Cartpole game with Gaussian noise on the image. The light blue arrows are added to show the direction in which the agent can move. Figure source: \citep{huang2021adarl}.
 \label{Figure: visual_cartpole}}
\end{figure} 
 
The original dataset \citep{huang2021adarl} introduces ``two change factors respectively for the state transition dynamics $\theta^{\text{dyn}}_k$: varying gravity and varying mass of the cart, and a change factor in the observation function $\theta^{\text{obs}}_k$ that is the image noise level.
Fig.~\ref{Figure: visual_cartpole} gives a visual example of Cartpole game, and the image with Gaussian noise. The images of the varying gravity and mass look exactly like the original image.
Specifically, in the gravity case, we consider source domains with gravity $g= \{5, 10, 20, 30, 40\}$. We take into account both interpolation (where the gravity in the target domain is in the support of that in source domains) with $g = \{15\}$,  and extrapolation (where it is out of the support w.r.t. the source domains) with $g = \{55\}$. 
Similarly, we consider source domains where the mass of the cart is $m = \{0.5, 1.5, 2.5, 3.5, 4.5\}$, while in target domains it is $m=\{1.0, 5.5\}$. In terms of changes on the observation function, we add Gaussian noise on the images with variance $\sigma = \{0.25, 0.75, 1.25, 1.75, 2.25 \}$ in source domains, and $\sigma = \{0.5, 2.75\}$ in target domains. The detailed settings in both source and target domains are in Table~\ref{Table: cartpole_setting}.''

 \begin{table}[h]
    \centering
    \begin{tabular}{c|c|c|c}
    \toprule
        & Gravity & Mass & Noise \\ \hline
    Source domains   &  $\{5, 10, 20, 30, 40\}$ & $\{0.5, 1.5, 2.5, 3.5, 4.5\}$ & $\{0.25, 0.75, 1.25, 1.75, 2.25\}$ \\ \hline
    Interpolation set  &  $\{15\}$ & $\{1.0\}$ & $\{0.5\}$ \\ \hline
    Extrapolation set  &  $\{55\}$ & $\{5.5\}$ & $\{2.75\}$ \\ \bottomrule
    \end{tabular}
    \caption{The settings of source and target domains for modified Cartpole experiments \citep{huang2021adarl}.}
    \label{Table: cartpole_setting}
\end{table}



\subsection{Mean Correlation Coefficient}
MCC is a standard metric for evaluating the recovery of latent factors in ICA literature. MCC first calculates the  absolute values of the correlation coefficient between every ground-truth factor against every estimated latent variable. Depending on whether componentwise invertible nonlinearities exist in the recovered factors, Pearson correlation coefficients or Spearman's rank correlation coefficients can be used. The possible permutation is adjusted by solving a linear sum assignment problem in polynomial time on the computed correlation matrix. 
\section{Implementation Details}

\subsection{Network Architecture}\label{ap:arch}

We summarize our network architecture below and describe it in detail in Table~\ref{tab:arch-details} and Table~\ref{tab:arch-cnndetails}.

\begin{table}[ht]
\caption{ Architecture details. BS: batch size, T: length of time series, i\_dim: input dimension, z\_dim: latent dimension, LeakyReLU: Leaky Rectified Linear Unit.}
\label{tab:arch-details}
\resizebox{\textwidth}{!}{%
\begin{tabular*}{1.05\textwidth}{@{\extracolsep{\fill}}|lll|}

\toprule
\textbf{Configuration} & \textbf{Description} &  \textbf{Output} \\
\toprule
\toprule
\textbf{1. MLP-Encoder} &  Encoder for Synthetic Data & \\
\toprule
Input: $\x_{1:T}$ & Observed time series & BS $\times$ T $\times$ i\_dim \\
Dense & 128 neurons, LeakyReLU & BS $\times$ T $\times$ 128\\
Dense & 128 neurons, LeakyReLU & BS $\times$ T $\times$ 128 \\
Dense & 128 neurons, LeakyReLU & BS $\times$ T $\times$ 128 \\
Dense & Temporal embeddings & BS $\times$ T $\times$ z\_dim \\
\toprule
\toprule
\textbf{2. MLP-Decoder} & Decoder for Synthetic Data & \\
\toprule
Input: $\hat{\z}_{1:T}$ & Sampled latent variables & BS $\times$ T $\times$ z\_dim \\
Dense & 128 neurons, LeakyReLU & BS $\times$ T $\times$ 128 \\
Dense & 128 neurons, LeakyReLU & BS $\times$ T $\times$ 128 \\
Dense & i\_dim neurons, reconstructed $\mathbf{\hat{x}}_{1:T}$ & BS $\times$ T $\times$ i\_dim \\
\toprule
\toprule
\textbf{5. Factorized Inference Network} & Bidirectional Inference Network & \\
\toprule
Input & Sequential embeddings & BS $\times$ T $\times$ z\_dim \\
Bottleneck & Compute mean and variance of posterior & $\mathbf{\mu}_{1:T}, \mathbf{\sigma}_{1:T}$ \\
Reparameterization & Sequential sampling & $\hat{\z}_{1:T}$ \\
\toprule
\toprule
\textbf{6. Modular Prior} & Nonlinear Transition Prior Network & \\
\toprule
Input & Sampled latent variable sequence $\hat{\z}_{1:T}$ & BS $\times$ T $\times$ z\_dim \\
InverseTransition & Compute estimated residuals $\hat{\epsilon}_{it}$ & BS $\times$ T $\times$ z\_dim \\
JacobianCompute & Compute $\log \left(\lvert \det \left(\mathbf{J}\right) \rvert \right)$ & BS\\
\bottomrule
\end{tabular*}
}
\end{table}

\begin{table}[ht]
\caption{ Architecture details on CNN encoder and decoder. BS: batch size, T: length of time series, h\_dim: hidden dimension, z\_dim: latent dimension, F: number of filters, (Leaky)ReLU: (Leaky) Rectified Linear Unit.}
\label{tab:arch-cnndetails}
\resizebox{\textwidth}{!}{%
\begin{tabular*}{1.05\textwidth}{@{\extracolsep{\fill}}|lll|}
\toprule
\textbf{Configuration} & \textbf{Description} &  \textbf{Output} \\
\toprule
\toprule
\textbf{3.1.1 CNN-Encoder} & Feature Extractor & \\
\toprule
Input: $\x_{1:T}$ & RGB video frames & BS $\times$ T $\times$ 3 $\times$ 64 $\times$ 64 \\
Conv2D & F: 32, BatchNorm2D, LeakyReLU & BS $\times$ T $\times$ 32 $\times$ 64 $\times$ 64 \\
Conv2D & F: 32, BatchNorm2D, LeakyReLU & BS $\times$ T $\times$ 32 $\times$ 32 $\times$ 32 \\
Conv2D & F: 32, BatchNorm2D, LeakyReLU & BS $\times$ T $\times$ 32 $\times$ 16 $\times$ 16 \\
Conv2D & F: 64, BatchNorm2D, LeakyReLU & BS $\times$ T $\times$ 64 $\times$ 8 $\times$ 8 \\
Conv2D & F: 64, BatchNorm2D, LeakyReLU & BS $\times$ T $\times$ 64 $\times$ 4 $\times$ 4 \\
Conv2D & F: 128, BatchNorm2D, LeakyReLU & BS $\times$ T $\times$ 128 $\times$ 1 $\times$ 1 \\
Dense & F: 2 * z\_dim = dimension of hidden embedding & BS $\times$ T $\times$ 2 * z\_dim \\
\toprule
\toprule
\textbf{4.1 CNN-Decoder} & Video Reconstruction & \\
\toprule
Input: $\z_{1:T}$ & Sampled latent variable sequence & BS $\times$ T $\times$ z\_dim \\
Dense & F: 128 , LeakyReLU & BS $\times$ T $\times$ 128 $\times$ 1 $\times$ 1\\
ConvTranspose2D & F: 64, BatchNorm2D, LeakyReLU & BS $\times$ T $\times$ 64 $\times$ 4 $\times$ 4 \\
ConvTranspose2D & F: 64, BatchNorm2D, LeakyReLU & BS $\times$ T $\times$ 64 $\times$ 8 $\times$ 8 \\
ConvTranspose2D & F: 32, BatchNorm2D, LeakyReLU & BS $\times$ T $\times$ 32 $\times$ 16 $\times$ 16 \\
ConvTranspose2D & F: 32, BatchNorm2D, LeakyReLU & BS $\times$ T $\times$ 32 $\times$ 32 $\times$ 32 \\
ConvTranspose2D & F: 32, BatchNorm2D, LeakyReLU & BS $\times$ T $\times$ 32 $\times$ 64 $\times$ 64 \\
ConvTranspose2D & F: 3, estimated scene $\mathbf{\hat{x}}_{1:T}$ & BS $\times$ T $\times$ 3 $\times$ 64 $\times$ 64 \\
\bottomrule
\end{tabular*}
}
\end{table}

\subsection{Hyperparameter and Training}

\paragraph{Hyperparameter Selection}

The hyperparameters of LiLY include $[\beta, \gamma]$, which are the weights of KLD terms for the time-lagged variables and the latent variables at the current time step, as well as the latent size $n$ and maximum time lag $L$. We use the ELBO loss on the validation dataset to select the best pair of $[\beta, \gamma, \sigma]$ because low ELBO loss always leads to high MCC. We always set a larger latent size than the true latent size. This is critical in video datasets because the image pixels contain more information than the annotated latent causal variables, and restricting the latent size will hurt the reconstruction performances. For the maximum time lag $L$, we set it by the rule of thumb. For instance, we use $L=2$ for temporal datasets with a latent physics process.

\paragraph{Training Details} 
 The models were implemented in \texttt{PyTorch} 1.8.1. The VAE network is trained using AdamW optimizer for a maximum of 50 epochs and early stops if the validation ELBO loss does not decrease for five epochs. A learning rate of 0.002 and a mini-batch size of 64 are used.  We have used three random seeds in each experiment and reported the mean performance with standard deviation averaged across random seeds. We have used several standard tricks to improve training stability: (1) we use a slightly larger latent size than the true latent size for real-world datasets in order to make sure the meaningful latent variables are among the recovered latent variables, and (2) we use AdamW optimizer as a regularizer to prevent training from being interrupted by overflow or underflow of variance terms of VAE.


\paragraph{Computing Hardware} 

We used a machine with the following CPU specifications: Intel(R) Core(TM) i7-7700K CPU @ 4.20GHz; 8 CPUs, four physical cores per CPU, a total of 32 logical CPU units. The machine has two GeForce GTX 1080 Ti  GPUs with 11GB GPU memory.
 
\paragraph{Reproducibility}

We've included the code for our framework and all experiments in the supplementary materials. We plan to release our code under the MIT License after the ICML paper review period.

\end{document}